\documentclass{article} 
\usepackage{iclr2026_conference,times}

\usepackage{amsmath,amsfonts,bm}

\def\eqref#1{equation~\ref{#1}}

\def\1{\bm{1}}

\DeclareMathAlphabet{\mathsfit}{\encodingdefault}{\sfdefault}{m}{sl}
\SetMathAlphabet{\mathsfit}{bold}{\encodingdefault}{\sfdefault}{bx}{n}

\usepackage{hyperref}
\usepackage{url}

\usepackage{amsfonts}
\usepackage{amssymb}
\usepackage{adjustbox}
\usepackage{booktabs}       
\usepackage{caption}
\usepackage{cleveref}
\usepackage[T1]{fontenc}    
\usepackage{graphicx}
\usepackage[utf8]{inputenc} 
\usepackage{listings}
\usepackage{makecell}
\usepackage{microtype}      
\usepackage{multirow}
\usepackage{nicefrac}       
\usepackage{pifont}
\usepackage{subcaption}
\usepackage{upquote}
\usepackage{wrapfig}
\usepackage{xcolor}         
\usepackage{enumitem}
\usepackage{textcomp}

\lstdefinestyle{mystyle}{
    basicstyle=\ttfamily\small,
    keywordstyle=\color{blue}\bfseries,
    stringstyle=\color{orange},
    commentstyle=\color{green!50!black}\itshape,
    breaklines=true,
    columns=fullflexible,
    showstringspaces=false,
    aboveskip=0pt,          
    belowskip=0pt,          
    lineskip=0pt            
}
\lstdefinestyle{prompt}{
  basicstyle=\ttfamily\small,
  frame=single,
  breaklines=true,
}

\lstdefinelanguage{yaml}{
  keywords={true,false,null,y,n},
  keywordstyle=\color{darkgray}\bfseries,
  basicstyle=\ttfamily\small,
  sensitive=false,
  comment=[l]{\#},
  morecomment=[s]{/*}{*/},
  commentstyle=\color{purple}\ttfamily,
  stringstyle=\color{blue}\ttfamily,
  moredelim=[l][\color{orange}]{\&},
  moredelim=[l][\color{magenta}]{*},
  moredelim=**[il][\color{red}{:}\color{blue}]{:},
  frame=single,
}

\def\eg{e.g.}

\def\ie{i.e.}

\definecolor{mygreen}{RGB}{0, 158, 115} 
\newcommand{\cmark}{\textcolor{mygreen}{\ding{51}}}  
\definecolor{myred}{RGB}{204, 121, 167} 
\newcommand{\xmark}{\textcolor{myred}{\ding{55}}}
\definecolor{myorange}{RGB}{230, 159, 0} 

\iclrfinalcopy 

\title{TAMA: Tool-Augmented Multimodal Agent \\for Procedural Activity Understanding}

\author{Kimihiro Hasegawa$^{1}$ \hspace{1em} Wiradee Imrattanatrai$^{2}$ \hspace{1em} Masaki Asada$^{2}$ \\
\textbf{Ken Fukuda}$^{2}$ \hspace{1em} \textbf{Teruko Mitamura}$^{1}$ \\
$^{1}$Language Technologies Institute, Carnegie Mellon University \\
$^{2}$National Institute of Advanced Industrial Science and Technology (AIST) \\
\texttt{kimihiro@cs.cmu.edu} 
}

\begin{document}

\maketitle

\begin{abstract}
Procedural activity assistants potentially support humans in a variety of settings, from our daily lives, e.g., cooking or assembling flat-pack furniture, to professional situations, e.g., manufacturing or biological experiments.
Despite its potential use cases, the system development tailored for such an assistant is still underexplored.
In this paper, we propose a novel framework, called TAMA, a Tool-Augmented Multimodal Agent, for procedural activity understanding.
TAMA enables interleaved multimodal reasoning by making use of multimedia-returning tools in a training-free setting. 
Our experimental result on the multimodal procedural QA dataset, ProMQA-Assembly, shows that our approach can improve the performance of vision-language models, especially GPT-5 and MiMo-VL. 
Furthermore, our ablation studies provide empirical support for the effectiveness of two features that characterize our framework, multimedia-returning tools and agentic flexible tool selection. 
We believe our proposed framework and experimental results facilitate the thinking with images paradigm for video and multimodal tasks, let alone the development of procedural activity assistants.\footnote{Code is available at \url{https://github.com/kimihiroh/tama}}
\end{abstract}

\section{Introduction}
\label{sec:introduction}

Procedural activities are ubiquitous, spanning our daily lives and professional settings, such as cooking~\citep{peddi-etal-NEURIPS2024-captaincook4d}, assembly~\citep{sener-cvpr2022-assembly101}, manufacturing~\citep{schoonbeek-wacv2024-industreal}, lab experiments~\citep{yagi-ijcv2025-finebio}, and medical practice~\citep{jang-arxiv2023-multimodal}, among others.
Assistant systems can democratize such activities by providing supportive guidance that makes them accessible to beginners.
Advances in large language models (LLMs) and vision-language models (VLMs) have significantly enhanced performance on existing video understanding benchmarks through improved pretraining and posttraining.
For further improvement, we combine the ideas of reasoning and agent to enable the ``thinking with images'' paradigm~\citep{su-arxiv2025-thinking} as an inference-time technique for procedural activity understanding.

Procedural activity understanding involves comprehending both the actual process, captured in the recording, and the expected process, described in textual or visual instructions, and aligning them to detect potential mismatches~\citep{hasegawa-etal-2025-promqa}.
This cross-modal alignment can be more tractable by decomposing the overall process into more manageable subtasks.
For instance, suppose one asks the following question while assembling a flat-pack furniture, ``Did I make any mistake before attaching this part?''
When humans approach this question, they typically examine the situation one by one. 
First, check the instructions to determine when and how the part is supposed to be attached. 
Next, they review the actions in the video to identify any misalignments, \eg, skipped steps or incorrect step orders. 
By repeating these steps as needed, they eventually either flag an error or conclude that no error exists and respond to the question. 

One naive, yet typical approach for such video-centric multimodal tasks with VLMs is to provide all information as one input, \ie, feed to a model the concatenation of a question, instructions, and sampled frames from a recording, and obtain a prediction in one inference.
This simple formulation aligns well with traditional workflow approaches that predefine information processing paths, \eg, keyframe selection that selects only informative frames, followed by answer prediction~\citep{ye-cvpr2025-rethinking}.
It also works well with recent techniques, like prompt engineering~\citep{liu-acm2023-prompt} or reasoning model~\cite{jaech-arxiv2024-openaio1}, both of which scale the inference time by outputting additional thought tokens, preceding its answer generation.
However, due to the nature of single-pass prediction, errors in beginning processes, \eg, frame selection, may be difficult to recover from, or managing context with multiple frames and many thought tokens poses long-dependency challenges for models~\citep{sun-etal-2025-mitigating-visual}. 

\begin{figure*}[t]
    \centering
    \includegraphics[width=\textwidth]{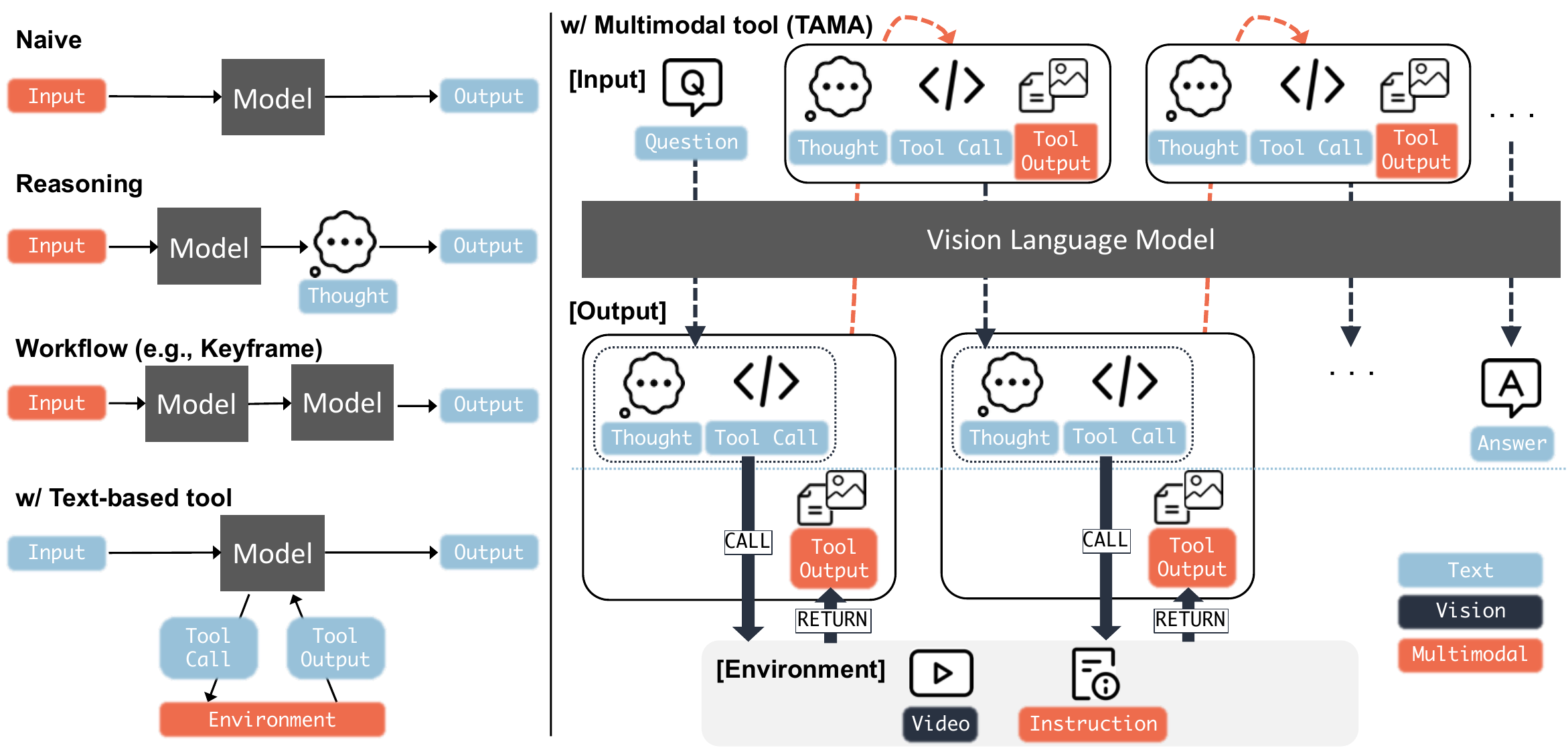}
    \caption{Left: Overview of existing approach. Right: Overview of our proposed approach, TAMA. Given a question as an initial input, a VLM-based agent generates its thought, followed by a tool call. Once a tool output is produced, the concatenation of the model output and the tool output is appended to the previous input to form the next input. Then, the model further generates either the next pair of a thought and tool call or an answer.}
    \label{fig:overview}
\end{figure*}

Another growing direction is an agentic approach~\citep{xi-sci25-rise}.
Compared to the single-pass approaches, which are typically implemented with a predefined, fixed-step workflow, a language model (LM) as an agent answers a question by making use of the predefined information processes (tools) flexibly, proactively, and iteratively.
Prior studies in video-centric tasks primarily design an agentic framework with text LLMs, which reason and decide actions, and semantic grounding tools, which provide textual conversion of visual information through captioning or OCR~\citep{wang-arxiv2024-videoagent}.
In contrast, humans would perform interleaved reasoning and visual comprehension by using perceptual exploration tools, \eg, fast-forwarding videos by time stamps, zooming into one frame, changing camera angles, etc. 
While the paradigm of interleaved textual and visual reasoning, \ie, thinking with images~\citep{openai-2024-think}, has been gaining attention on image-centric tasks, its application to video-centric tasks, especially in training-free settings, is yet underexplored.
Inspired by this gap, we pose the following research question: \textit{Can VLMs make use of interleaved multimodal reasoning by using perceptual exploration tools to better perform video-centric multimodal tasks?}

In this work, we propose a novel training-free agentic framework, \textbf{TAMA} (\textbf{T}ool-\textbf{A}ugmented \textbf{M}ultimodal \textbf{A}gent), that enables interleaved multimodal reasoning by multimedia-returning tool use.
\autoref{fig:overview} illustrates the overview of our proposed framework.
A VLM-based agent orchestrates tools that return either images or text to perceptually explore the current situation with its reasoning capability in an interleaved manner (\S~\ref{sec:approach}). 
We experiment with our framework in a training-free setting, where VLMs are given only task and tool information as a prompt to see if current VLMs can make use of the tools out of the box. 
Our experimental results with both proprietary and open-weight models on ProMQA-Assembly~\citep{hasegawa-etal-2025-promqa-assembly} reveal that our framework can further elicit the performance for some models, \ie, GPT-5~\citep{openai-2025-gpt5} and MiMo-VL~\citep{coreteam-arxiv2025-mimovl}, although the performance change varies, as sometimes performance degrades under our framework, \ie, Qwen2.5-VL~\citep{bai-arxiv2025-qwen25} and InternVL3~\citep{zhu-arxiv2025-internvl3}.
Yet, the result suggests that our framework can potentially bring out the models' capability in a zero-shot manner for video-centric multimodal tasks, considering that, while arguable, most of the tools from our experiments are likely to be unseen during training.
As a behavioral analysis, we examined the tool-use patterns, aiming to provide potential reasons for performance discrepancies (\S~\ref{sec:experiment}).
Furthermore, to provide the empirical evidence of our framework's efficacy, we conducted the ablation studies, w.r.t the modality of tool outputs, the flexibility of the tool selection, and the impact of frame sampling (\S~\ref{sec:ablation-study}).

In short, our contributions are threefold:
(1) We propose a training-free agentic framework for interleaved multimodal reasoning with tools.
(2) Our experiment shows that our framework can potentially improve the performance of VLMs.
(3) Our ablation studies support that multimedia-return tools and agentic tool use are beneficial. 
We believe that our work stimulates research on the thinking with images paradigm for video understanding tasks, thus, more capable procedural activity assistants that benefit human society. 
\section{Related work}
\label{sec:related_work}
Our work is inspired by reasoning VLMs, video agents, and procedural activity understanding. 

\subsection{Vision-Language Model}
Vision-language models, which process visual and textual information together, have rapidly progressed over the past few years. 
Strong proprietary models are mostly VLMs by default~\citep{openai-2025-gpt5, anthropic-2025-claude4, google-2025-gemini25}, and an increasing amount of competitive open/open-weights models have been released in the community~\citep{bai-arxiv2025-qwen25, zhu-arxiv2025-internvl3, coreteam-arxiv2025-mimovl}.
On top of the popular prompt techniques~\citep{wei-arxiv2022-chain,wang-iclr2023-self}, reasoning models are becoming dominant in public benchmarks~\citep{jaech-arxiv2024-openaio1, guo-arxiv2025-deepseek}.
While a reasoning paradigm is primarily on text, its variant, ``thinking with images''~\citep{su-arxiv2025-thinking}, has also been gaining attention.
This paradigm introduces visual information into its textual thought process in an interleaved manner by making use of external tools~\citep{openai-2024-think, hu-neurips2024-visual} or by using a native multimodal model that has the capability of synthesizing images as well~\citep{team-arxiv2024-chameleon}.
Our work aligns with the former tool-driven thinking with images paradigm, specifically for video understanding tasks. 

\subsection{Video Agent}
In video understanding studies, a traditional workflow system~\citep{anthropic-2025-building}, where a model processes data based on a fixed predefined order, has played a major role and is still competitive, due to its customizability, \eg, Socratic model~\citep{zeng-arxiv2022-socratic} or keyframe selection approaches~\citep{ye-cvpr2025-rethinking,arnab-arxiv2025-temporal}.
In parallel to the general progress of VLMs and their agentic capability, agentic approaches are getting more attention in video tasks as well~\citep{xi-sci25-rise}.
An agent, typically a language model, flexibly and proactively selects an action based on tool descriptions and its thought process to understand the situation and answer a question.
While prior agentic systems in video understanding tasks show their effectiveness, in most cases, their tools are for semantic grounding, \ie, converting images into text, with only a textual thought process~\citep{wang-arxiv2024-videoagent,tian-arxiv2025-egor1}.
In contrast, our work features perceptual exploration tools, which help an agent to explore visually~\citep{wu-cvpr2024-v,zhang-arxiv2025-chain} and form interleaved multimodal reasoning.
One concurrent work by \citet{zhang-2025-thinking} also proposes to use frame sampling as a tool; however, their with-training and single-tool setup differs from our training-free and multi-tool setup.

\subsection{Procedural Activity Understanding}
Procedural activity exists everywhere, where assistants can support users from their ego- and exocentric viewpoints by aligning observed actions in recordings with the expected actions in instructions.
Due to the ubiquitous demands, prior studies have covered diverse domains: 
cooking~\citep{stein-ubicomp2013-salad,peddi-etal-NEURIPS2024-captaincook4d, lee-cvpr2024-egoper}, assembly~\citep{ben-shabat-wacv2021=ikea-asm, jang-iccv2019-epic-tent}, manufacturing~\citep{ragusa-wacv2021-meccano, wang-iccv-2023-holoassist, schoonbeek-wacv2024-industreal}, lab experiments~\citep{yagi-ijcv2025-finebio}, and medical practice~\citep{beyer-2016-virtual,jang-arxiv2023-multimodal}, among others~\citep{haneji-arxiv2024-egooops}.
While classification tasks are popular in those studies, some work explores other task formulations to facilitate the development of systems with more human-friendly and detailed responses.
For instance, the ProMQA series proposes multimodal QA datasets on procedural activities, i.e., cooking and assembly~\citep{hasegawa-etal-2025-promqa, hasegawa-etal-2025-promqa-assembly}. 
In this work, we adopt ProMQA-Assembly as our evaluation dataset, considering the instruction variety, i.e., target assembly image, in addition to both textual and image instructions (Example in~\autoref{tab:analysis-example}).
We leave it to future work to apply our method to ProMQA(-cooking) or other datasets.

\section{Approach}
\label{sec:approach}

TAMA is a training-free agentic framework that enables interleaved multimodal reasoning by tool use. 
Before introducing our approach, we first define our target task, followed by the existing approaches.
All approaches, including TAMA, are illustrated on~\autoref{fig:overview}.

\subsection{Task Formulation}
\label{ssec:task-formulation}
Our target task is multimodal question answering, specifically for understanding procedural activities. 
The input consists of: (1) a user's question in text, (2) a video recording of the activity up to the point when the question is asked, and (3) instructions provided in both image and text formats, including a target assembly image. 
The output is a textual answer.

\subsection{Existing Approach}
\label{ssec:existing-approach}
\paragraph{Naive and Reasoning}
One prevalent approach with VLMs feeds the concatenation of sampled frames from a video, instructions, and a question into models (\textit{naive})~\citep{fu-cvpr2025-videomme}. 
On top of this naive approach, prompt techniques or reasoning models are used to further enhance the performance (\textit{reasoning}). 
While simple, depending on a model's valid context length, a model may not keep attending enough attention to initial frames in its decoding time~\citep{sun-etal-2025-mitigating-visual}.

\paragraph{Workflow}
Most traditional studies can be categorized into \textit{workflow}, where processes, \eg, LLMs and tools, follow a predefined sequential path.
For instance, keyframe selection approaches can be seen as workflow systems when you treat the first stage of keyframe selection and the second stage of answer generation as two fixed-order processes/tools~\citep{arnab-arxiv2025-temporal}.
While customizable, since the process path needs to be predefined, careful path design would be required (\S~\ref{ssec:agentic-vs-workflow}).

\paragraph{Agent with text-returning tool}
Arguably, due to the success of text LMs, this has been the major approach for existing agentic work for video understanding tasks:
An agent, \ie, a text-only LM, devises an answer in response to a query/question by flexibly making use of tools that return text.
When a tool is invoked, it accesses the environment for a textual instruction or a video file. 
When the target is text, the information is passed through the tool and returned to the agent.
When the target is visual content, a tool, typically VLMs or task-specific models, performs semantic grounding by converting it into text, \eg, captioning, and returns it to the agent.
While this approach can benefit from the evolving agentic capability of text-only LMs, vision-to-text conversion can be an information bottleneck, which may impair performance (\S~\ref{ssec:text-vs-multimedia}).

\subsection{Ours: TAMA}
\label{ssec:ours}

\begin{table*}[!t]
\centering
\caption{Our tool set.}
\begin{adjustbox}{width=\linewidth}
\begin{tabular}{c l c}
    \toprule
    Function & Description & Example \\
    \midrule
    
    Sample frame &
    \makecell[l]{Return sampled frames in the specified range \\at equal intervals from the specified angle of the camera.} &
    \begin{minipage}{0.6\linewidth}
        \begin{lstlisting}[language=Python]
sample_frame(
    start='0:10', end='0:20', angle='center'
)
        \end{lstlisting}
    \end{minipage} \\ \midrule
    
    Zoom in & 
    \makecell[l]{Return the specified frame's cropped image based \\on the specified normalized bounding box.} & 
    \begin{minipage}{0.6\linewidth}
        \begin{lstlisting}[language=Python]
zoom_in(
    frame_id=106, 
    bounding_box=[0.3, 0.4, 0.7, 0.8]
)
        \end{lstlisting}
    \end{minipage} \\ \midrule
    
    Check instruction & 
    \makecell[l]{Return an instruction in either text (DOT format\footnote{\url{https://en.wikipedia.org/wiki/DOT_(graph_description_language)}}) \\or image (directed acyclic graph).} &
    \begin{minipage}{0.6\linewidth}
        \begin{lstlisting}[language=Python]
check_instruction(mode='text')
        \end{lstlisting}
    \end{minipage} \\ \midrule
    
    Check final picture & 
    Return the target assembly image with parts. &
    \begin{minipage}{0.6\linewidth}
        \begin{lstlisting}[language=Python]
check_final_picture()
        \end{lstlisting}
    \end{minipage} \\
    
    \bottomrule
\end{tabular}
\end{adjustbox}
\label{tab:tool-set}
\end{table*}

Our approach employs a VLM rather than a text-only LM as its agent and relies on multimedia-returning tools that return information in original modalities, \ie, text remains text and images remain images.
Existing agent frameworks have proposed to integrate VLMs for video understanding tasks, yet mainly as tools, rather than agents~\citep{yang-arxiv2023-mmreact, tian-arxiv2025-egor1}. 
Motivated by the success of GUI agents~\citep{zhang-arxiv2024-gui}, we propose to use VLMs as agents for video understanding tasks so that an agent can reason and call tools based on original multimodal information. 
To leverage the capability of VLM-based agents, we define four tools, as summarized in Table~\ref{tab:tool-set}. 
\texttt{sample\_frame} and \texttt{zoom\_in} enable an agent to explore a video at different granularities.
\texttt{check\_instruction} and \texttt{check\_final\_picture} help an agent to access manuals in different modalities.
Essentially, the tools are defined so that models can explore information perceptually, rather than ground visual information in text, to prevent any information loss during information conversion.
Tools are all implemented as Python functions that access local files.
We explore this framework in a training-free setting to investigate current VLMs' zero-shot capability.
As illustrated in~\autoref{fig:overview}, we feed a prompt with a question (and tool information) to a model and generate a tool call with a thought process. 
Once we obtain a tool output by executing the tool locally, we append both the model output and the tool output to the previous input, which is again fed to a model. 

\section{Experiment}
\label{sec:experiment}

\begin{table*}[!t]
\centering
\caption{Result.}
\fontsize{8}{9}\selectfont
\begin{tabular}{l c c c c}
    \toprule
    Model & Naive & Reasoning & TCoT & TAMA (ours) \\
    \midrule
     GPT-5 mini & 58.1 & 56.9 & 58.8 & \textbf{63.7} \\
     GPT-5 & 58.7& 60.0& 57.9& \textbf{67.0}\\
     Claude 4 Sonnet & 46.4& \textbf{56.8}& 52.0& 55.6\\
     Gemini 2.5 Flash& 41.6& 48.8& \textbf{54.9}& 52.4\\
     Qwen2.5-VL 32B & 44.0& \textbf{44.6}& 40.8& 44.0\\
     InternVL3 38B& \textbf{50.5}& 48.2& 48.5& 46.3\\
     MiMo-VL 7B& 33.1& 46.4& 46.8& \textbf{49.6}\\
    \bottomrule
\end{tabular}
\label{tab:result}
\end{table*}

We compare TAMA against existing approaches on a multimodal QA task to verify its effectiveness.

\subsection{Baseline Approach}
\label{ssec:baseline}
We first compare our framework with three baseline approaches: naive, reasoning, and workflow.
For the naive and reasoning approaches, we feed the concatenation of sampled frames, an instruction (text), a target assembly image, and a question as one input, and obtain an answer, preceded by a thought process for the reasoning. 
For workflow, we experiment with Temporal Chain-of-Thought (TCoT)~\citep{arnab-arxiv2025-temporal}, a two-stage approach, where VLMs select keyframes based on each question, and the same model answers it based on the selected frames.
We chose TCoT as our baseline because it is also a training-free approach. 
As for the text tool-based agentic approach, we conduct an ablation study to compare text tools and multimedia tools in \S~\ref{ssec:text-vs-multimedia}.
As all approaches, including ours, are model-agnostic, we apply these approaches to the following models.

\subsection{Experimental Setup}
\label{ssec:experimental-setup}
In our experiment, we include both proprietary and open-weight models.
For proprietary models, we chose GPT-5, GPT-5 mini, Claude 4 Sonnet, and Gemini 2.5 Flash, based on their performance on public benchmarks and costs.
For open-weight models, among VLMs, we selected three models based on their reported capabilities on agentic benchmarks and also computational demands: Qwen2.5-VL 32B, InternVL3 38B, and MiMo-VL 7B.
To achieve TAMA's interleaved thought process, we use either reasoning mode for proprietary models and MiMo-VL 7B, or ReAct-style prompting~\citep{yao-iclr2023-react} with zero-shot CoT~\citep{kojima-neurips2022-large} for Qwen2.5-VL 32B and InternVL3 38B.
We format our iterative thought-call-return process in a similar way to multi-turn conversations, where we set minimum and maximum turns as hyperparameters. 
In case a model outputs an answer too quickly or too late, we include a cut-in message to encourage the model to think more or answer in the next turn.
All experiments are done without any in-context examples, \ie, zero-shot inference.

As our evaluation dataset, we use ProMQA-Assembly, a multimodal QA dataset for procedural activity understanding, which has a unique setting of including video recording, instructions, a target assembly image, and a question as input.
Following the prior work, we adopt the LLM-as-a-judge for assessing the quality of predictions.
A judge model outputs the score, 0 (incorrect), 1 (partially correct), or 2 (correct), and we take the average with scaling to 0 to 100 by multiplying by 50. 
All numbers are reported by a single run of experiments.
More details are available in Appendix~\ref{appx:exp-details}.

\begin{table*}[t]
\centering
\fontsize{8}{9}\selectfont
\caption{
    Example with an instruction image (top left), target assembly image with parts (top right), sampled frames from a recording (middle), and a pair of a question and ground-truth answers, followed by GPT-5's responses from each approach.
}
\setlength\tabcolsep{2pt}
\begin{tabular}{l p{12cm}}
    \toprule
    \multicolumn{2}{c}{\includegraphics[width=\linewidth]{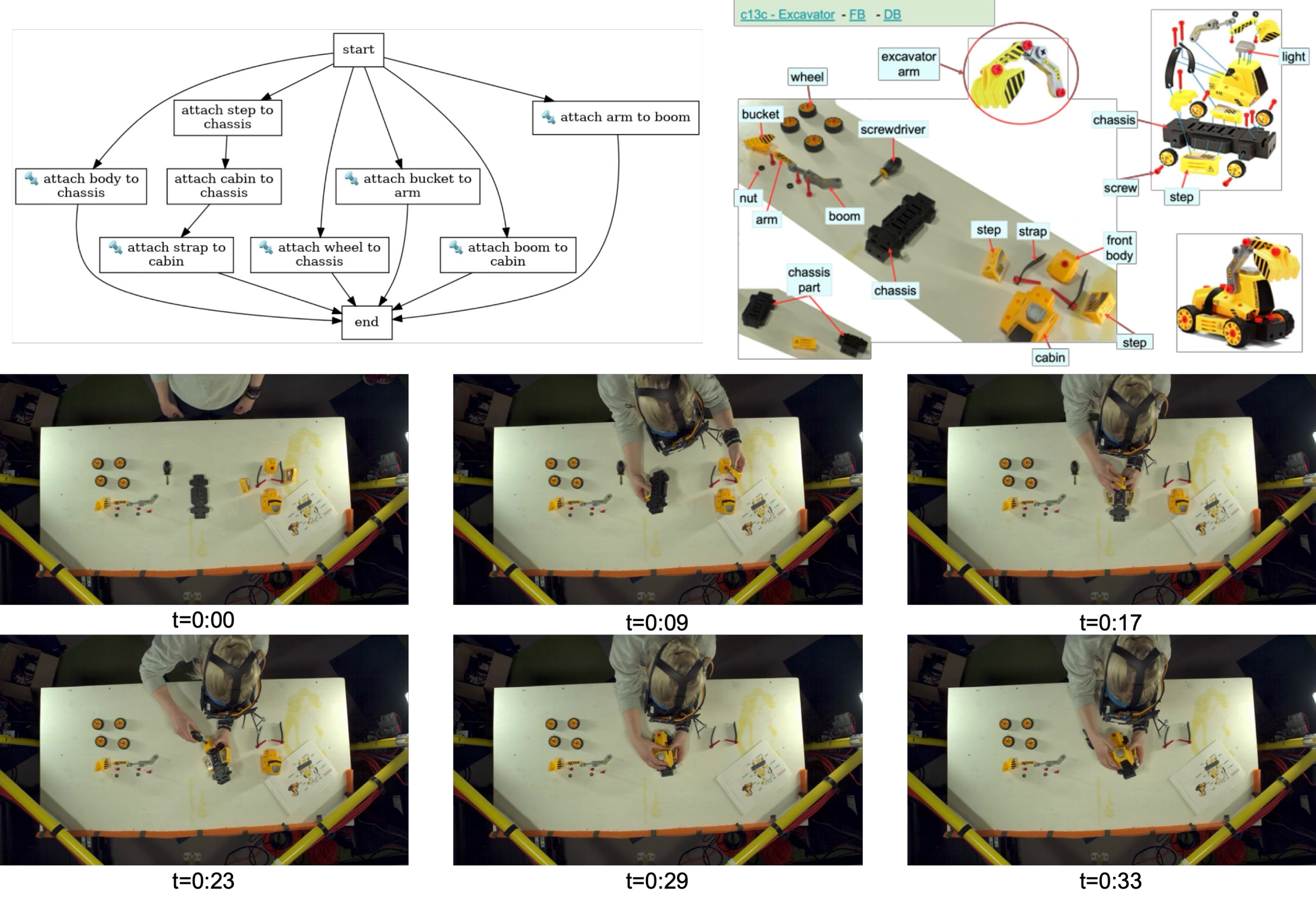}}\\
    Question: & What should I have done differently when attaching the body to the chassis? \\
    Ground Truth: & Nothing, everything is going great! / There was nothing you needed to do differently.\\
    \midrule
    Naive: & \xmark~You should have secured the body to the chassis with a screw instead of just snapping it on.\\
    Reasoning: & ... Maybe they should have used a different orientation for the screw, ...\\
    & \xmark~You should have secured the body to the chassis with a screw. \\
    TCoT: & \{"selected\_frames": [...], "justifications": "These frames show the actual step of attaching the yellow body onto the black chassis, including how it’s oriented and aligned. ..."\}, ... The instructions clearly state to use a screw for this step. ... the answer points towards using a screw ... \\
    & \xmark~Secure the body to the chassis with a screw—use the screwdriver to fasten it instead of just placing it on. \\
    TAMA: & ... I think I can call the functions to get the needed details. <tool=sample\_frame> <tool\_output=frames> There could be several issues: wrong orientation, incorrect screws, or missing steps like inserting the window piece first. ... It might be helpful to check the instructions ... <tool=check\_instruction> <tool\_output=instruction> ... call check\_final\_picture to understand the correct orientation. ...<tool=check\_finale\_picture> <tool\_output=image> ... The answer likely involves checking the orientation and ensuring the correct number of screws align with the holes. ... Using zoom tools on specific frames might help clarify things ... <tool=zoom\_in> <tool\_output=image> ... I’ll call for a zoom-in again ... <tool=zoom\_in> <tool\_output=image> ... I want to zoom in ... I'll request a different bounding box ...<tool=zoom\_in> <tool\_output=image> ... Despite some ambiguity from the video, ... they did it correctly ... \\
    & \cmark~Nothing—you aligned the front body correctly on the chassis and secured it with a screw as the instructions require. \\
    \bottomrule
\end{tabular}
\label{tab:analysis-example}
\end{table*}

\subsection{Result and Discussion}
\label{ssec:result}

\begin{table}[!t]
\centering
\caption{Analysis of TAMA.}
\fontsize{8}{9}\selectfont
\begin{tabular}{l c c c c c c c c}
    \toprule
    \multirow{2}{*}[-0.5ex]{Model} & \multirow{2}{*}[-0.5ex]{\makecell{\#frames \\ (avg./median)}} & \multicolumn{5}{c}{Tool Frequency per Question} & \multicolumn{2}{c}{\#turn} \\
    \cmidrule(lr){3-7} \cmidrule(lr){8-9}
    & & sample & zoom & inst. & pic. & total & 1st ans. & total \\
    \midrule
    GPT-5 mini & 20.8 / 20.0 & 1.2 & 0.4 & 1.1 & 1.0 & 3.7 & 3.1 & 8.0 \\
    GPT-5 & 24.2 / 21.0 & 1.2 & 1.8 & 1.2 & 1.0 & 5.2 & 4.0 & 8.8 \\
    Claude 4 Sonnet & 31.8 / 26.0 & 1.9 & 0.9 & 1.6 & 1.1 & 5.4 & 3.2 & 9.0 \\
    Gemini 2.5 Flash & 12.7 / 7.0 & 1.1 & 0.4 & 1.5 & 0.6 & 3.6 & 5.1 & 7.3 \\
    Qwen2.5-VL 32B & 22.5 / 26.0 & 1.2 & 0.2 & 0.9 & 0.8 & 3.1 & 4.7 & 9.3 \\
    InternVL3 38B & 18.0 / 14.0 & 1.6 & 0.3 & 1.1 & 1.0 & 3.9 & 4.3 & 9.9 \\
    MiMo-VL 7B & 14.1 / 11.0 & 1.3 & 0.2 & 0.9 & 0.8 & 3.2 & 3.9 & 9.2 \\
    \bottomrule
\end{tabular}
\label{tab:num-frame}
\end{table}
\begin{figure}[!t]
    \centering
    
    \begin{subfigure}[t]{0.4\textwidth}
        \centering
        \includegraphics[width=\textwidth]{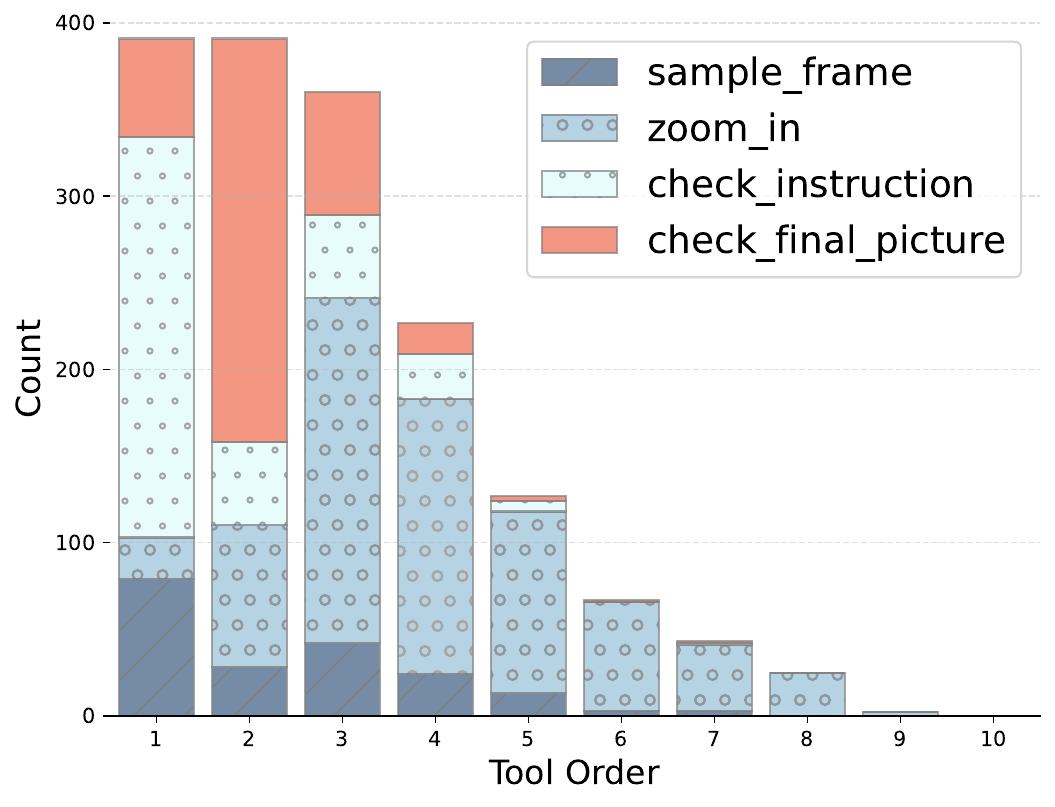}
        \caption{GPT-5}
    \end{subfigure}
    \begin{subfigure}[t]{0.4\textwidth}
        \centering
        \includegraphics[width=\textwidth]{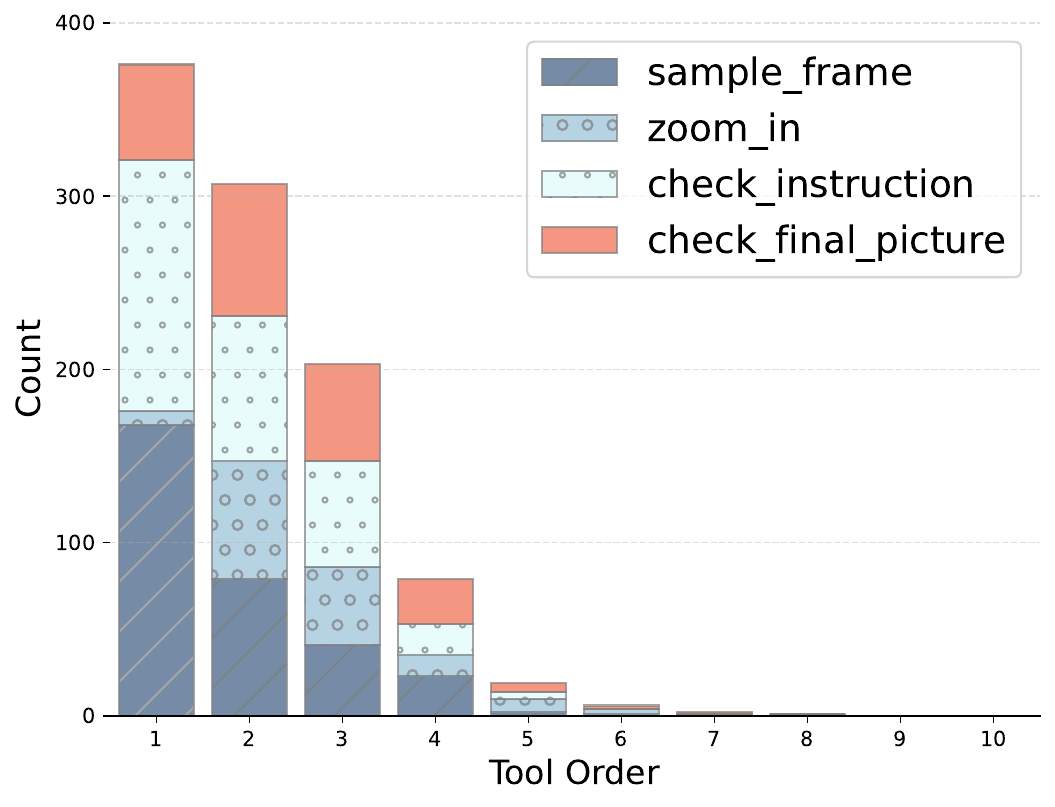}
        \caption{Gemini 2.5 Flash}
    \end{subfigure}
    
    \vspace{0.5em}
    
    \begin{subfigure}[t]{0.4\textwidth}
        \centering
        \includegraphics[width=\textwidth]{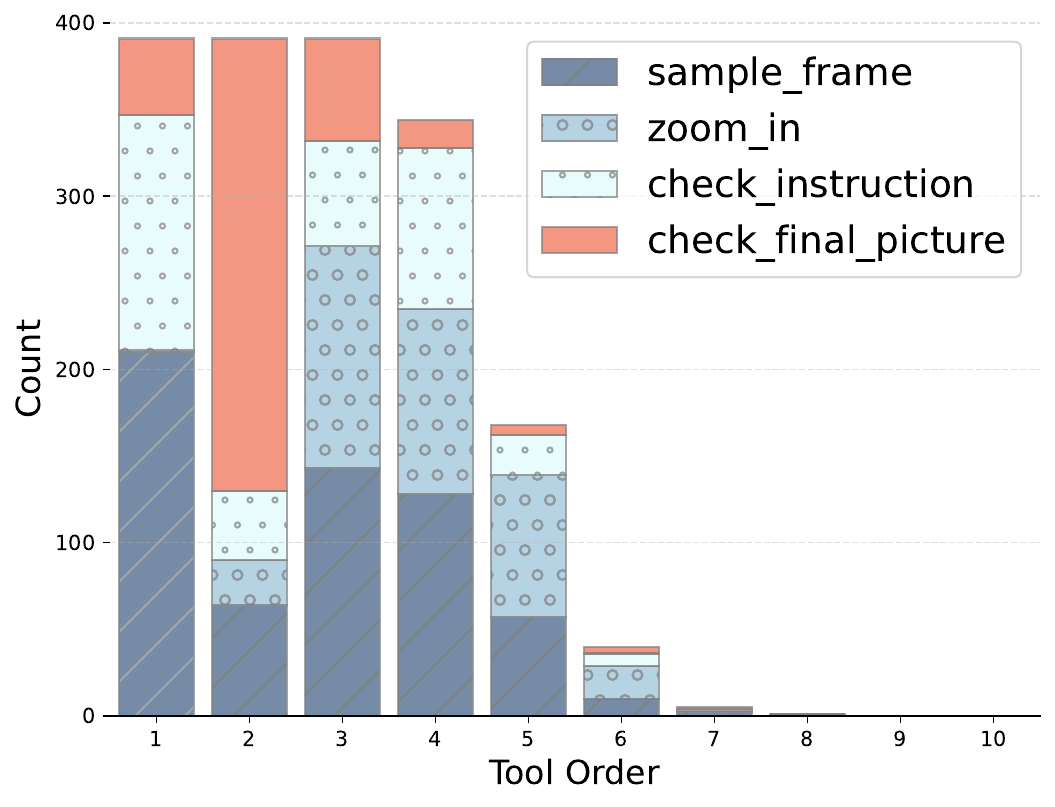}
        \caption{Claude 4 Sonnet}
    \end{subfigure}
    \begin{subfigure}[t]{0.4\textwidth}
        \centering
        \includegraphics[width=\textwidth]{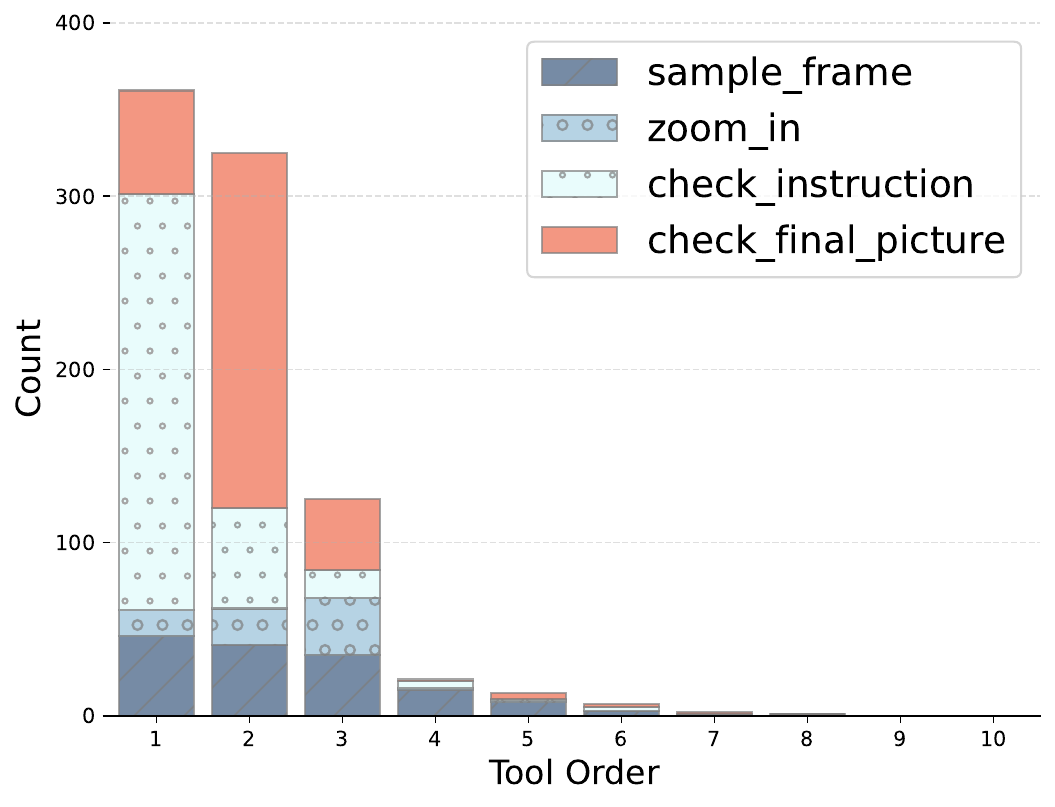}
        \caption{MiMo-VL 7B}
    \end{subfigure}
    
    \caption{Tool usage pattern.}
    \label{fig:analysis-pattern}
\end{figure}

\autoref{tab:result} shows the experimental result. 
Among all the combinations, GPT-5 with our framework exhibits the best performance. 
For GPT-5, GPT-5 mini, and MiMo-VL 7B, TAMA outperforms other approaches for each model, \eg, 14.1\% improvement from the naive approach to TAMA by GPT-5. 
Gemini 2.5 Flash with TAMA shows superior performance over naive and reasoning approaches, but lags behind TCoT.
Claude 4 Sonnet prefers our framework over naive and TCoT, but its text-only reasoning process shows slightly better performance than ours.
For Qwen2.5-VL 32B and InternVL3 38B, neither TAMA nor TCoT outperforms the naive or reasoning approaches. 

To better understand model differences, we investigated several aspects of each model's output: 
the total number of sampled frames, tool usage frequency, and the number of turns for initial and final answers per question (\autoref{tab:num-frame}). 
We also examined tool usage patterns (\autoref{fig:analysis-pattern}).
Gemini 2.5 Flash sampled a substantially smaller number of frames, which can be a potential reason for its less performant result with TAMA (See \S~\ref{ssec:presample} for our empirical support).
As the Gemini API documentation\footnote{\url{https://ai.google.dev/gemini-api/docs/video-understanding}} describes that it can specify points in a video by a timestamp, the model is expected to be familiar with timestamps. 
Thus, since the model calls \texttt{sampel\_frame} in a similar frequency to other models, it may tend to select fewer frames, as reported in the TCoT paper. 
In contrast, Claude 4 Sonnet sampled the largest number of frames among all models, even though the model does not benefit from our framework.
This suggests that the number of frames itself does not correlate with the effectiveness of our framework. 
GPT-5 and Claude 4 Sonnet call tools more frequently than others, where GPT-5 notably prefers the zoom-in tool, as highlighted in the tool pattern figures.
This indicates that, in conjunction with its superior performance, GPT-5 may be specifically trained for the thinking with images paradigm with similar tools, and the capability may be transferable to video understanding tasks under our framework. 
\autoref{tab:analysis-example} shows one set of example outputs from GPT-5. 
The model uses \texttt{zoom\_in} tools in the latter half of the process to be more certain of its answer.

Qwen2.5-VL 32B and InternVL3 38B show similar characteristics to GPT-5 mini, in terms of the number of frames and tool frequency (\autoref{fig:appx-analysis-pattern} in Appendix).
However, during our manual inspection, we noticed that these open-weight models sometimes failed to follow the intended ReAct-style prompting. 
While we expected an interleaved thought process, the models occasionally refused to output any reasoning and instead produced only tool calls one after another.
This suggests that these models would require additional tuning to be applied in our framework.
MiMo-VL 7B does not show any particular uniqueness in its tool frequency or pattern, while it is the only open-weight model that benefits from our framework.
Based on the claims in the MiMo-VL paper and its result (naive $<$ reasoning $<$ TAMA) in our experiment, one can guess that the capability of textual reasoning may be related to interleaved multimodal reasoning. 
However, as the result of Claude 4 Sonnet may refute (naive $<$ TAMA $\leq$ reasoning), further investigation would be needed to understand what training contributes to interleaved multimodal reasoning, and we leave it for future work.

\section{Ablation Study}
\label{sec:ablation-study}

Our proposed framework, TAMA, distinguishes itself from prior work in two aspects: multimedia-return tools and agentic, flexible tool selections. 
To further understand their effects, we conducted the two ablation studies.
In addition, we experimented with one heuristic strategy, presampling, inspired by the undersampling behavior of Gemini 2.5 Flash. 

\subsection{Text-return Tool vs Multimedia-return Tool}
\label{ssec:text-vs-multimedia}

\begin{table}[!t]
\centering
\begin{minipage}[c]{0.48\textwidth}
\centering
\caption{Perf. w/ Text vs Multimedia tool.}
\fontsize{8}{9}\selectfont
\begin{tabular}{l c c}
    \toprule
    Model & Text & Multi \\
    \midrule
     GPT-5 mini & 59.0 & 63.7 \\
     Gemini 2.5 Flash & 48.2 & 52.4 \\
     Qwen2.5-VL 32B & 39.0 & 42.1 \\
     MiMo-VL 7B & 50.9 & 49.6 \\
    \bottomrule
\end{tabular}
\label{tab:text-vs-multi}
\end{minipage}
\hfill
\begin{minipage}[c]{0.48\textwidth}
\centering
\caption{Perf. w/ and w/o presample.}
\fontsize{8}{9}\selectfont
\begin{tabular}{l c c}
    \toprule
    Model & TAMA & \makecell[c]{TAMA \\w/ presample} \\
    \midrule
     GPT-5 mini & 63.7 & 63.2 \\
     Gemini 2.5 Flash & 52.4 & 55.0 \\
     Qwen2.5-VL 32B & 44.0 & 49.0 \\
     InternVL3 38B & 46.3 & 46.7\\
     MiMo-VL 7B & 49.6 & 49.1 \\
    \bottomrule
\end{tabular}
\label{tab:ablation-presample}
\end{minipage}
\end{table}
\begin{figure*}[t]
    \centering
    \includegraphics[width=0.8\textwidth]{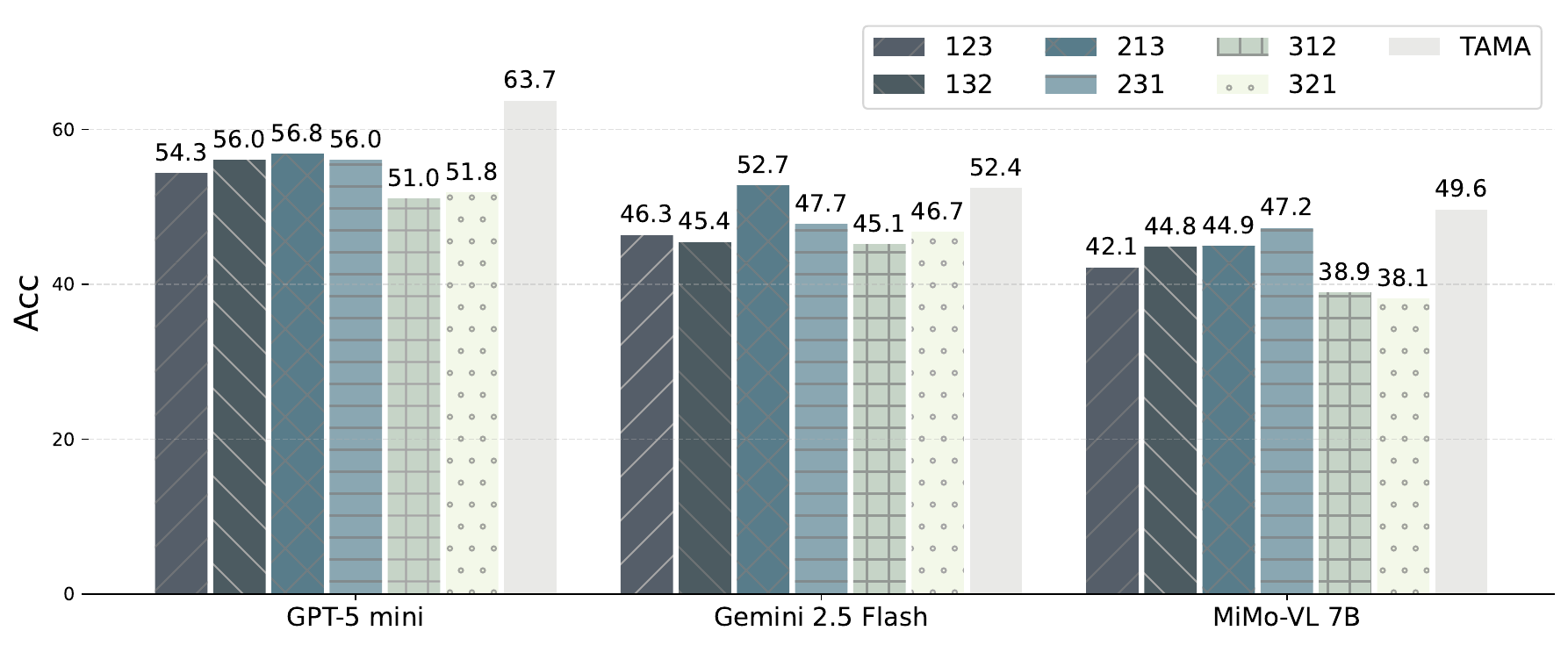}
    \caption{Performance of workflow vs agentic approach (TAMA). Each number represents one tool operation in the workflow approach: ``1'' is the uniform sampling, ``2'' is the instruction check, and ``3'' is the target assembly image check, and each digit sequence defines the execution order of the tools.}
    \label{fig:ablation-order}
\end{figure*}

The first characteristic lies in tools capable of returning multimedia outputs.
Given a tool call, our tools can return either text or images, contrary to the text-returning tools.
As mentioned in \S~\ref{ssec:existing-approach}, we conducted a controlled experiment by defining a semantic-grounding version of our perceptual exploration tools.
Specifically, image-returning tools are instead returning captions of images, where captions are obtained by prompting the same model as its agent model.
To isolate the effect of agent models, we use VLMs for both text-returning and multimedia-returning tools, instead of text LMs, which are typical for agents with text-returning tools.
GPT-5 mini, Gemini 2.5 Flash, Qwen2.5-VL 32B, and MiMo-VL 7B are used in this experiment.
According to the result in~\autoref{tab:text-vs-multi}, GPT-5 mini, Gemini 2.5 Flash, and Qwen2.5-VL 32B with multimedia-returning tools outperform those with text-returning tools, while MiMo-VL 7B prefers text-returning tools. 
One possible reason for the MiMo-VL's preference may stem from its video re-captioning pipeline for pretraining, where they produced dense, fine-grained captions for each video. 
However, overall, our experiment shows a positive impact of multimedia-returning tools.

\subsection{Workflow vs Agentic Tool Use}
\label{ssec:agentic-vs-workflow}

Secondly, we investigate the effect of its proactive and flexible tool selection.
Specifically, we compared TAMA with a fixed-order workflow approach. 
We selected the following three tools with fixing arguments: namely, \texttt{sample\_frame} with uniform sampling from each recording, \texttt{check\_instruction} with text mode, and \texttt{check\_final\_picture}.
The outputs of these tools are fed to a model sequentially, while the model is prompted to output only its thought process without any tool calls. 
Once all information is given, a model is instructed to produce an answer.
In this experiment, we included all the permutations of these three operations (6 orders in total) using GPT-5 mini, Gemini 2.5 Flash, and MiMo-VL 7B.
\autoref{fig:ablation-order} summarizes the result.
We found that all permutations of the workflow approach degraded the performance, regardless of tool orders, except for one combination.
When Gemini 2.5 Flash received information in the order of textual instructions, sampled frames, and the target assembly image, it performed comparably to TAMA.
These results demonstrate the superior performance and cost-effectiveness of the agentic approach compared to the workflow-based method.
Although the workflow approach can be tuned to match the agentic approach's performance, the agentic approach demonstrates superior usability. 
It achieves comparable or better performance without tuning by flexibly selecting appropriate tools and execution orders for each question, making it more efficient and user-friendly.

\subsection{Presampling}
\label{ssec:presample}

As we found in our investigation (\S~\ref{ssec:result}), some models, \ie, Gemini 2.5 Flash and MiMo-VL 7B, tend to select fewer frames than others. 
\citet{arnab-arxiv2025-temporal} addresses this point by compensating with uniformly sampled frames in their TCoT approach.
Inspired by this, we also hypothesize that feeding additional frames may benefit those models. 
Specifically, we append the uniformly sampled frames from each recording to the initial prompt, which consists of a question and task information.
This can be thought of as a hybrid approach of workflow and agentic framework. 
We conducted this experiment to get a better sense of which tool selection capabilities would be beneficial to incorporate into future training for video understanding tasks, with a specific focus on sampling. 
We primarily targeted Gemini 2.5 Flash, InternVL3 38B, and MiMo-VL 7B, as they had fewer sampled frames. 
We also included GPT-5 mini and Qwen2.5-VL 32B for comparison.
As shown in~\autoref{tab:ablation-presample}, Gemini 2.5 Flash gains the benefit from this presampled strategy, while the performance of InternVL3 38B and MiMo-VL 7B did not change.
Contrary to our expectation, Qwen2.5-VL 32B improves its performance with this strategy, although the number of its sampled frames is around the average of other models. 
While some models have already shown their capability of making use of our framework, this presampling experiment implies that additional training with respect to sampling may benefit these models.

\section{Conclusion}
\label{sec:conclusion}
In this work, we propose a novel training-free agentic framework, TAMA, to enable interleaved multimodal reasoning with tool use. 
Our experimental result shows that our framework for the thinking with images paradigm improves the performance of models such as GPT-5, GPT-5 mini, and MiMo-VL 7B.
While some other models, Gemini 2.5 Flash and Qwen2.5-VL 32B, show their potential with the hybrid approach with presampling, the other models, \eg, Claude 4 Sonnet or InternVL3 38B, do not gain benefits, arguably because they are not familiar with an interleaved reasoning process or zero-shot use of our tools.
Yet, together with the ablation study results on multimedia-returning tools and agentic tool selection, our work provides empirical support for our zero-shot, agentic prompting technique in a multi-turn setting.
We believe that our work can facilitate the research on the perceptual exploration tools and interleaved multimodal reasoning for video understanding tasks, let alone the development of procedural activity assistants that benefit human society.

\section*{Ethics Statement}
Our work does not introduce any training data, which may introduce additional biases or harmful content to VLMs. 
However, the negative contents inherent in VLMs from pretraining or posttraining may emerge within our framework.
If our framework is to be deployed for production, rigorous evaluation against biases, fairness, privacy, jailbreak, etc, needs to be performed on top of our performance-focused evaluation, including the thought process.

\section*{Reproducibility Statement}
We provide the general description of our proposed approach in \S~\ref{ssec:ours} and the experimental setup in \S~\ref{ssec:experimental-setup}, which is further detailed in Appendix~\ref{appx:exp-details}.
We also provide the prompt templates for our experiment in Appendix~\ref{appx:prompt}.
Furthermore, we attach the anonymized code used in our experiments as a supplemental material. 
\section*{Acknowledgment}
This work is partially supported by Programs for Bridging the gap between R\&D and the IDeal society (society 5.0) and Generating Economic and social value (BRIDGE) / Practical Global Research in the AI $\times$ Robotics Services, implemented by the Cabinet Office, Government of Japan.

\clearpage

\bibliography{iclr2026_conference}

\begin{thebibliography}{47}
\providecommand{\natexlab}[1]{#1}
\providecommand{\url}[1]{\texttt{#1}}
\expandafter\ifx\csname urlstyle\endcsname\relax
  \providecommand{\doi}[1]{doi: #1}\else
  \providecommand{\doi}{doi: \begingroup \urlstyle{rm}\Url}\fi

\bibitem[Anthropic(2025{\natexlab{a}})]{anthropic-2025-building}
Anthropic.
\newblock Building effective agents, 2025{\natexlab{a}}.
\newblock URL \url{https://www.anthropic.com/engineering/building-effective-agents}.

\bibitem[Anthropic(2025{\natexlab{b}})]{anthropic-2025-claude4}
Anthropic.
\newblock Introducing claude 4, 2025{\natexlab{b}}.
\newblock URL \url{https://www.anthropic.com/news/claude-4}.

\bibitem[Arnab et~al.(2025)Arnab, Iscen, Caron, Fathi, and Schmid]{arnab-arxiv2025-temporal}
Anurag Arnab, Ahmet Iscen, Mathilde Caron, Alireza Fathi, and Cordelia Schmid.
\newblock Temporal chain of thought: Long-video understanding by thinking in frames.
\newblock \emph{arXiv preprint arXiv:2507.02001}, 2025.

\bibitem[Bai et~al.(2025)Bai, Chen, Liu, Wang, Ge, Song, Dang, Wang, Wang, Tang, et~al.]{bai-arxiv2025-qwen25}
Shuai Bai, Keqin Chen, Xuejing Liu, Jialin Wang, Wenbin Ge, Sibo Song, Kai Dang, Peng Wang, Shijie Wang, Jun Tang, et~al.
\newblock Qwen2. 5-vl technical report.
\newblock \emph{arXiv preprint arXiv:2502.13923}, 2025.

\bibitem[Ben-Shabat et~al.(2021)Ben-Shabat, Yu, Saleh, Campbell, Rodriguez-Opazo, Li, and Gould]{ben-shabat-wacv2021=ikea-asm}
Yizhak Ben-Shabat, Xin Yu, Fatemeh Saleh, Dylan Campbell, Cristian Rodriguez-Opazo, Hongdong Li, and Stephen Gould.
\newblock The ikea asm dataset: Understanding people assembling furniture through actions, objects and pose.
\newblock In \emph{Proceedings of the IEEE/CVF Winter Conference on Applications of Computer Vision (WACV)}, pp.\  847--859, January 2021.

\bibitem[Beyer-Berjot et~al.(2016)Beyer-Berjot, Berdah, Hashimoto, Darzi, and Aggarwal]{beyer-2016-virtual}
Laura Beyer-Berjot, St{\'e}phane Berdah, Daniel~A Hashimoto, Ara Darzi, and Rajesh Aggarwal.
\newblock A virtual reality training curriculum for laparoscopic colorectal surgery.
\newblock \emph{Journal of surgical education}, 73\penalty0 (6):\penalty0 932--941, 2016.

\bibitem[Fu et~al.(2025)Fu, Dai, Luo, Li, Ren, Zhang, Wang, Zhou, Shen, Zhang, et~al.]{fu-cvpr2025-videomme}
Chaoyou Fu, Yuhan Dai, Yongdong Luo, Lei Li, Shuhuai Ren, Renrui Zhang, Zihan Wang, Chenyu Zhou, Yunhang Shen, Mengdan Zhang, et~al.
\newblock Video-mme: The first-ever comprehensive evaluation benchmark of multi-modal llms in video analysis.
\newblock In \emph{Proceedings of the Computer Vision and Pattern Recognition Conference}, pp.\  24108--24118, 2025.

\bibitem[Google(2025)]{google-2025-gemini25}
Google.
\newblock Gemini 2.5: Our most intelligent ai model, 2025.
\newblock URL \url{https://blog.google/technology/google-deepmind/gemini-model-thinking-updates-march-2025/#gemini-2-5-thinking}.

\bibitem[Guo et~al.(2025)Guo, Yang, Zhang, Song, Zhang, Xu, Zhu, Ma, Wang, Bi, et~al.]{guo-arxiv2025-deepseek}
Daya Guo, Dejian Yang, Haowei Zhang, Junxiao Song, Ruoyu Zhang, Runxin Xu, Qihao Zhu, Shirong Ma, Peiyi Wang, Xiao Bi, et~al.
\newblock Deepseek-r1: Incentivizing reasoning capability in llms via reinforcement learning.
\newblock \emph{arXiv preprint arXiv:2501.12948}, 2025.

\bibitem[Haneji et~al.(2024)Haneji, Nishimura, Kameko, Shirai, Yoshida, Kajimura, Yamamoto, Cui, Nishimoto, and Mori]{haneji-arxiv2024-egooops}
Yuto Haneji, Taichi Nishimura, Hirotaka Kameko, Keisuke Shirai, Tomoya Yoshida, Keiya Kajimura, Koki Yamamoto, Taiyu Cui, Tomohiro Nishimoto, and Shinsuke Mori.
\newblock Egooops: A dataset for mistake action detection from egocentric videos with procedural texts, 2024.

\bibitem[Hasegawa et~al.(2025{\natexlab{a}})Hasegawa, Imrattanatrai, Asada, Holm, Wang, Zhou, Fukuda, and Mitamura]{hasegawa-etal-2025-promqa-assembly}
Kimihiro Hasegawa, Wiradee Imrattanatrai, Masaki Asada, Susan Holm, Yuran Wang, Vincent Zhou, Ken Fukuda, and Teruko Mitamura.
\newblock Promqa-assembly: Multimodal procedural qa dataset on assembly.
\newblock \emph{arXiv preprint arXiv:2509.02949}, 2025{\natexlab{a}}.

\bibitem[Hasegawa et~al.(2025{\natexlab{b}})Hasegawa, Imrattanatrai, Cheng, Asada, Holm, Wang, Fukuda, and Mitamura]{hasegawa-etal-2025-promqa}
Kimihiro Hasegawa, Wiradee Imrattanatrai, Zhi-Qi Cheng, Masaki Asada, Susan Holm, Yuran Wang, Ken Fukuda, and Teruko Mitamura.
\newblock {P}ro{MQA}: Question answering dataset for multimodal procedural activity understanding.
\newblock In Luis Chiruzzo, Alan Ritter, and Lu~Wang (eds.), \emph{Proceedings of the 2025 Conference of the Nations of the Americas Chapter of the Association for Computational Linguistics: Human Language Technologies (Volume 1: Long Papers)}, pp.\  11598--11617, Albuquerque, New Mexico, April 2025{\natexlab{b}}. Association for Computational Linguistics.
\newblock ISBN 979-8-89176-189-6.

\bibitem[Hong et~al.(2025)Hong, Yu, Gu, Wang, Gan, Tang, Cheng, Qi, Ji, Pan, et~al.]{hong-arxiv2025-glm}
Wenyi Hong, Wenmeng Yu, Xiaotao Gu, Guo Wang, Guobing Gan, Haomiao Tang, Jiale Cheng, Ji~Qi, Junhui Ji, Lihang Pan, et~al.
\newblock Glm-4.1 v-thinking: Towards versatile multimodal reasoning with scalable reinforcement learning.
\newblock \emph{arXiv e-prints}, pp.\  arXiv--2507, 2025.

\bibitem[Hu et~al.(2024)Hu, Shi, Fu, Roth, Ostendorf, Zettlemoyer, Smith, and Krishna]{hu-neurips2024-visual}
Yushi Hu, Weijia Shi, Xingyu Fu, Dan Roth, Mari Ostendorf, Luke Zettlemoyer, Noah~A. Smith, and Ranjay Krishna.
\newblock Visual sketchpad: Sketching as a visual chain of thought for multimodal language models.
\newblock In \emph{The Thirty-eighth Annual Conference on Neural Information Processing Systems}, 2024.

\bibitem[Jaech et~al.(2024)Jaech, Kalai, Lerer, Richardson, El-Kishky, Low, Helyar, Madry, Beutel, Carney, et~al.]{jaech-arxiv2024-openaio1}
Aaron Jaech, Adam Kalai, Adam Lerer, Adam Richardson, Ahmed El-Kishky, Aiden Low, Alec Helyar, Aleksander Madry, Alex Beutel, Alex Carney, et~al.
\newblock Openai o1 system card.
\newblock \emph{arXiv preprint arXiv:2412.16720}, 2024.

\bibitem[Jang et~al.(2019)Jang, Sullivan, Ludwig, Gilchrist, Damen, and Mayol-Cuevas]{jang-iccv2019-epic-tent}
Youngkyoon Jang, Brian Sullivan, Casimir Ludwig, Iain Gilchrist, Dima Damen, and Walterio Mayol-Cuevas.
\newblock Epic-tent: An egocentric video dataset for camping tent assembly.
\newblock In \emph{Proceedings of the IEEE/CVF International Conference on Computer Vision (ICCV) Workshops}, Oct 2019.

\bibitem[Jang et~al.(2023)Jang, Sohn, Logeswaran, Luo, Lee, and Lee]{jang-arxiv2023-multimodal}
Yunseok Jang, Sungryull Sohn, Lajanugen Logeswaran, Tiange Luo, Moontae Lee, and Honglak Lee.
\newblock Multimodal subtask graph generation from instructional videos.
\newblock \emph{arXiv preprint arXiv:2302.08672}, 2023.

\bibitem[Kojima et~al.(2022)Kojima, Gu, Reid, Matsuo, and Iwasawa]{kojima-neurips2022-large}
Takeshi Kojima, Shixiang~Shane Gu, Machel Reid, Yutaka Matsuo, and Yusuke Iwasawa.
\newblock Large language models are zero-shot reasoners.
\newblock In \emph{Proceedings of the 36th International Conference on Neural Information Processing Systems}, NIPS '22, Red Hook, NY, USA, 2022. Curran Associates Inc.
\newblock ISBN 9781713871088.

\bibitem[Lee et~al.(2024)Lee, Lu, Zhang, Hoai, and Elhamifar]{lee-cvpr2024-egoper}
Shih-Po Lee, Zijia Lu, Zekun Zhang, Minh Hoai, and Ehsan Elhamifar.
\newblock Error detection in egocentric procedural task videos.
\newblock In \emph{Proceedings of the IEEE/CVF Conference on Computer Vision and Pattern Recognition (CVPR)}, pp.\  18655--18666, June 2024.

\bibitem[Liu et~al.(2023)Liu, Yuan, Fu, Jiang, Hayashi, and Neubig]{liu-acm2023-prompt}
Pengfei Liu, Weizhe Yuan, Jinlan Fu, Zhengbao Jiang, Hiroaki Hayashi, and Graham Neubig.
\newblock Pre-train, prompt, and predict: A systematic survey of prompting methods in natural language processing.
\newblock \emph{ACM Comput. Surv.}, 55\penalty0 (9), January 2023.
\newblock ISSN 0360-0300.
\newblock \doi{10.1145/3560815}.

\bibitem[OpenAI(2025{\natexlab{a}})]{openai-2024-think}
OpenAI.
\newblock Thinking with images, 2025{\natexlab{a}}.
\newblock URL \url{https://openai.com/index/thinking-with-images/}.

\bibitem[OpenAI(2025{\natexlab{b}})]{openai-2025-gpt5}
OpenAI.
\newblock Introducing gpt-5, 2025{\natexlab{b}}.
\newblock URL \url{https://openai.com/index/introducing-gpt-5/}.

\bibitem[Peddi et~al.(2024)Peddi, Arya, Challa, Pallapothula, Vyas, Gouripeddi, Zhang, Wang, Komaragiri, Ragan, Ruozzi, Xiang, and Gogate]{peddi-etal-NEURIPS2024-captaincook4d}
Rohith Peddi, Shivvrat Arya, Bharath Challa, Likhitha Pallapothula, Akshay Vyas, Bhavya Gouripeddi, Qifan Zhang, Jikai Wang, Vasundhara Komaragiri, Eric Ragan, Nicholas Ruozzi, Yu~Xiang, and Vibhav Gogate.
\newblock Captaincook4d: A dataset for understanding errors in procedural activities.
\newblock In A.~Globerson, L.~Mackey, D.~Belgrave, A.~Fan, U.~Paquet, J.~Tomczak, and C.~Zhang (eds.), \emph{Advances in Neural Information Processing Systems}, volume~37, pp.\  135626--135679. Curran Associates, Inc., 2024.

\bibitem[Ragusa et~al.(2021)Ragusa, Furnari, Livatino, and Farinella]{ragusa-wacv2021-meccano}
Francesco Ragusa, Antonino Furnari, Salvatore Livatino, and Giovanni~Maria Farinella.
\newblock The meccano dataset: Understanding human-object interactions from egocentric videos in an industrial-like domain.
\newblock In \emph{Proceedings of the IEEE/CVF Winter Conference on Applications of Computer Vision (WACV)}, pp.\  1569--1578, January 2021.

\bibitem[Schoonbeek et~al.(2024)Schoonbeek, Houben, Onvlee, van~der Sommen, et~al.]{schoonbeek-wacv2024-industreal}
Tim~J Schoonbeek, Tim Houben, Hans Onvlee, Fons van~der Sommen, et~al.
\newblock Industreal: A dataset for procedure step recognition handling execution errors in egocentric videos in an industrial-like setting.
\newblock In \emph{Proceedings of the IEEE/CVF Winter Conference on Applications of Computer Vision}, pp.\  4365--4374, 2024.

\bibitem[Sener et~al.(2022)Sener, Chatterjee, Shelepov, He, Singhania, Wang, and Yao]{sener-cvpr2022-assembly101}
Fadime Sener, Dibyadip Chatterjee, Daniel Shelepov, Kun He, Dipika Singhania, Robert Wang, and Angela Yao.
\newblock Assembly101: A large-scale multi-view video dataset for understanding procedural activities.
\newblock In \emph{Proceedings of the IEEE/CVF Conference on Computer Vision and Pattern Recognition (CVPR)}, pp.\  21096--21106, June 2022.

\bibitem[Stein \& McKenna(2013)Stein and McKenna]{stein-ubicomp2013-salad}
Sebastian Stein and Stephen~J. McKenna.
\newblock Combining embedded accelerometers with computer vision for recognizing food preparation activities.
\newblock In \emph{Proceedings of the 2013 ACM International Joint Conference on Pervasive and Ubiquitous Computing}, UbiComp '13, pp.\  729–738, New York, NY, USA, 2013. Association for Computing Machinery.
\newblock ISBN 9781450317702.
\newblock \doi{10.1145/2493432.2493482}.

\bibitem[Su et~al.(2025)Su, Xia, Guo, Liu, Ma, Qu, Liu, Li, Zeng, Yang, et~al.]{su-arxiv2025-thinking}
Zhaochen Su, Peng Xia, Hangyu Guo, Zhenhua Liu, Yan Ma, Xiaoye Qu, Jiaqi Liu, Yanshu Li, Kaide Zeng, Zhengyuan Yang, et~al.
\newblock Thinking with images for multimodal reasoning: Foundations, methods, and future frontiers.
\newblock \emph{arXiv preprint arXiv:2506.23918}, 2025.

\bibitem[Sun et~al.(2025)Sun, Sun, Peng, and Ye]{sun-etal-2025-mitigating-visual}
Hai-Long Sun, Zhun Sun, Houwen Peng, and Han-Jia Ye.
\newblock Mitigating visual forgetting via take-along visual conditioning for multi-modal long {C}o{T} reasoning.
\newblock In Wanxiang Che, Joyce Nabende, Ekaterina Shutova, and Mohammad~Taher Pilehvar (eds.), \emph{Proceedings of the 63rd Annual Meeting of the Association for Computational Linguistics (Volume 1: Long Papers)}, pp.\  5158--5171, Vienna, Austria, July 2025. Association for Computational Linguistics.
\newblock ISBN 979-8-89176-251-0.

\bibitem[Team(2024)]{team-arxiv2024-chameleon}
Chameleon Team.
\newblock Chameleon: Mixed-modal early-fusion foundation models.
\newblock \emph{arXiv preprint arXiv:2405.09818}, 2024.

\bibitem[Tian et~al.(2025)Tian, Wang, Guo, Wu, Dong, Wang, Yang, Zhang, Zhu, and Liu]{tian-arxiv2025-egor1}
Shulin Tian, Ruiqi Wang, Hongming Guo, Penghao Wu, Yuhao Dong, Xiuying Wang, Jingkang Yang, Hao Zhang, Hongyuan Zhu, and Ziwei Liu.
\newblock Ego-r1: Chain-of-tool-thought for ultra-long egocentric video reasoning.
\newblock \emph{arXiv preprint arXiv:2506.13654}, 2025.

\bibitem[Wang et~al.(2024)Wang, Zhang, Zohar, and Yeung-Levy]{wang-arxiv2024-videoagent}
Xiaohan Wang, Yuhui Zhang, Orr Zohar, and Serena Yeung-Levy.
\newblock Videoagent: Long-form video understanding with large language model as agent.
\newblock In \emph{European Conference on Computer Vision}, pp.\  58--76. Springer, 2024.

\bibitem[Wang et~al.(2023{\natexlab{a}})Wang, Kwon, Rad, Pan, Chakraborty, Andrist, Bohus, Feniello, Tekin, Frujeri, Joshi, and Pollefeys]{wang-iccv-2023-holoassist}
Xin Wang, Taein Kwon, Mahdi Rad, Bowen Pan, Ishani Chakraborty, Sean Andrist, Dan Bohus, Ashley Feniello, Bugra Tekin, Felipe~Vieira Frujeri, Neel Joshi, and Marc Pollefeys.
\newblock Holoassist: an egocentric human interaction dataset for interactive ai assistants in the real world.
\newblock In \emph{Proceedings of the IEEE/CVF International Conference on Computer Vision (ICCV)}, pp.\  20270--20281, October 2023{\natexlab{a}}.

\bibitem[Wang et~al.(2023{\natexlab{b}})Wang, Wei, Schuurmans, Le, Chi, Narang, Chowdhery, and Zhou]{wang-iclr2023-self}
Xuezhi Wang, Jason Wei, Dale Schuurmans, Quoc~V Le, Ed~H. Chi, Sharan Narang, Aakanksha Chowdhery, and Denny Zhou.
\newblock Self-consistency improves chain of thought reasoning in language models.
\newblock In \emph{The Eleventh International Conference on Learning Representations}, 2023{\natexlab{b}}.

\bibitem[Wei et~al.(2022)Wei, Wang, Schuurmans, Bosma, Xia, Chi, Le, Zhou, et~al.]{wei-arxiv2022-chain}
Jason Wei, Xuezhi Wang, Dale Schuurmans, Maarten Bosma, Fei Xia, Ed~Chi, Quoc~V Le, Denny Zhou, et~al.
\newblock Chain-of-thought prompting elicits reasoning in large language models.
\newblock \emph{Advances in neural information processing systems}, 35:\penalty0 24824--24837, 2022.

\bibitem[Wu \& Xie(2024)Wu and Xie]{wu-cvpr2024-v}
Penghao Wu and Saining Xie.
\newblock V?: Guided visual search as a core mechanism in multimodal llms.
\newblock In \emph{Proceedings of the IEEE/CVF Conference on Computer Vision and Pattern Recognition (CVPR)}, pp.\  13084--13094, June 2024.

\bibitem[Xi et~al.(2025)Xi, Chen, Guo, He, Ding, Hong, Zhang, Wang, Jin, Zhou, Zheng, Fan, Wang, Xiong, Zhou, Wang, Jiang, Zou, Liu, Yin, Dou, Weng, Qin, Zheng, Qiu, Huang, Zhang, and Gui]{xi-sci25-rise}
Zhiheng Xi, Wenxiang Chen, Xin Guo, Wei He, Yiwen Ding, Boyang Hong, Ming Zhang, Junzhe Wang, Senjie Jin, Enyu Zhou, Rui Zheng, Xiaoran Fan, Xiao Wang, Limao Xiong, Yuhao Zhou, Weiran Wang, Changhao Jiang, Yicheng Zou, Xiangyang Liu, Zhangyue Yin, Shihan Dou, Rongxiang Weng, Wenjuan Qin, Yongyan Zheng, Xipeng Qiu, Xuanjing Huang, Qi~Zhang, and Tao Gui.
\newblock The rise and potential of large language model based agents: a survey.
\newblock \emph{Sci. China Inf. Sci.}, 68\penalty0 (2), 2025.

\bibitem[Xiaomi(2025)]{coreteam-arxiv2025-mimovl}
LLM-Core-Team Xiaomi.
\newblock Mimo-vl technical report, 2025.

\bibitem[Yagi et~al.(2025)Yagi, Ohashi, Huang, Furuta, Adachi, Mitsuyama, and Sato]{yagi-ijcv2025-finebio}
Takuma Yagi, Misaki Ohashi, Yifei Huang, Ryosuke Furuta, Shungo Adachi, Toutai Mitsuyama, and Yoichi Sato.
\newblock Finebio: A fine-grained video dataset of biological experiments with hierarchical annotation.
\newblock \emph{International Journal of Computer Vision}, pp.\  1--16, 2025.

\bibitem[Yang et~al.(2023)Yang, Li, Wang, Lin, Azarnasab, Ahmed, Liu, Liu, Zeng, and Wang]{yang-arxiv2023-mmreact}
Zhengyuan Yang, Linjie Li, Jianfeng Wang, Kevin Lin, Ehsan Azarnasab, Faisal Ahmed, Zicheng Liu, Ce~Liu, Michael Zeng, and Lijuan Wang.
\newblock Mm-react: Prompting chatgpt for multimodal reasoning and action.
\newblock \emph{arXiv preprint arXiv:2303.11381}, 2023.

\bibitem[Yao et~al.(2023)Yao, Zhao, Yu, Du, Shafran, Narasimhan, and Cao]{yao-iclr2023-react}
Shunyu Yao, Jeffrey Zhao, Dian Yu, Nan Du, Izhak Shafran, Karthik Narasimhan, and Yuan Cao.
\newblock {ReAct}: Synergizing reasoning and acting in language models.
\newblock In \emph{International Conference on Learning Representations (ICLR)}, 2023.

\bibitem[Ye et~al.(2025)Ye, Wang, Sun, Chandrasegaran, Durante, Eyzaguirre, Bisk, Niebles, Adeli, Fei-Fei, Wu, and Li]{ye-cvpr2025-rethinking}
Jinhui Ye, Zihan Wang, Haosen Sun, Keshigeyan Chandrasegaran, Zane Durante, Cristobal Eyzaguirre, Yonatan Bisk, Juan~Carlos Niebles, Ehsan Adeli, Li~Fei-Fei, Jiajun Wu, and Manling Li.
\newblock Re-thinking temporal search for long-form video understanding.
\newblock In \emph{Proceedings of the IEEE/CVF Conference on Computer Vision and Pattern Recognition (CVPR)}, pp.\  8579--8591, June 2025.

\bibitem[Zeng et~al.(2022)Zeng, Attarian, Ichter, Choromanski, Wong, Welker, Tombari, Purohit, Ryoo, Sindhwani, et~al.]{zeng-arxiv2022-socratic}
Andy Zeng, Maria Attarian, Brian Ichter, Krzysztof Choromanski, Adrian Wong, Stefan Welker, Federico Tombari, Aveek Purohit, Michael Ryoo, Vikas Sindhwani, et~al.
\newblock Socratic models: Composing zero-shot multimodal reasoning with language.
\newblock \emph{arXiv preprint arXiv:2204.00598}, 2022.

\bibitem[Zhang et~al.(2024)Zhang, He, Qian, Li, Li, Qin, Kang, Ma, Liu, Lin, et~al.]{zhang-arxiv2024-gui}
Chaoyun Zhang, Shilin He, Jiaxu Qian, Bowen Li, Liqun Li, Si~Qin, Yu~Kang, Minghua Ma, Guyue Liu, Qingwei Lin, et~al.
\newblock Large language model-brained gui agents: A survey.
\newblock \emph{arXiv preprint arXiv:2411.18279}, 2024.

\bibitem[Zhang et~al.(2025{\natexlab{a}})Zhang, Gu, Li, Ma, Bai, Zhang, Zhang, Zhou, He, and Tang]{zhang-2025-thinking}
Haoji Zhang, Xin Gu, Jiawen Li, Chixiang Ma, Sule Bai, Chubin Zhang, Bowen Zhang, Zhichao Zhou, Dongliang He, and Yansong Tang.
\newblock Thinking with videos: Multimodal tool-augmented reinforcement learning for long video reasoning.
\newblock \emph{arXiv preprint arXiv:2508.04416}, 2025{\natexlab{a}}.

\bibitem[Zhang et~al.(2025{\natexlab{b}})Zhang, Gao, Zhang, Li, Zhang, Liu, Yuan, Wu, Jia, Zhu, et~al.]{zhang-arxiv2025-chain}
Xintong Zhang, Zhi Gao, Bofei Zhang, Pengxiang Li, Xiaowen Zhang, Yang Liu, Tao Yuan, Yuwei Wu, Yunde Jia, Song-Chun Zhu, et~al.
\newblock Chain-of-focus: Adaptive visual search and zooming for multimodal reasoning via rl.
\newblock \emph{arXiv preprint arXiv:2505.15436}, 2025{\natexlab{b}}.

\bibitem[Zhu et~al.(2025)Zhu, Wang, Chen, Liu, Ye, Gu, Tian, Duan, Su, Shao, et~al.]{zhu-arxiv2025-internvl3}
Jinguo Zhu, Weiyun Wang, Zhe Chen, Zhaoyang Liu, Shenglong Ye, Lixin Gu, Hao Tian, Yuchen Duan, Weijie Su, Jie Shao, et~al.
\newblock Internvl3: Exploring advanced training and test-time recipes for open-source multimodal models.
\newblock \emph{arXiv preprint arXiv:2504.10479}, 2025.

\end{thebibliography}
\bibliographystyle{iclr2026_conference}

\clearpage
\appendix
\section{Experiment Details}
\label{appx:exp-details}

We share the further details of our experiments in this section.

\subsection{TCoT Implementation}
\label{appx:implemtation-details-tcot}
TCoT consists of two stages: the first stage is frame selection, and the second is answer generation. 
We used the dynamic-segment TCoT, where each input video is split into a fixed number of $l$ segments and each segment is fed to a model that generates the indices of frames for the second answer generation stage. 
Given the maximum number of frames, $k$, in each inference, if more than $k$ frames exist in one segment, $k$ frames are sampled from each segment. 
Once frames are selected from each segment, they are concatenated to form an input for answer generation. 
When the number of frames in this concatenation is more than $m$, $m$ frames are uniformly sampled.
In addition to the selected frames in the first stage, they add uniformly sampled $u$ frames for temporal coverage. 
Thus, the input of the second stage consists of at most $m+u$ frames, which is fed to a model with a question and instructions to generate an answer. 
Hyperparameters are set as follows: $l = 4$, $k = 32$, $m=48$, and $u=16$. 
Prompt templates are available in~\autoref{fig:prompt-tcot-frame-selection} and~\ref{fig:prompt-tcot-answer-generation}.

\subsection{TAMA Implementation}
\label{appx:implemtation-details-tama}

As described in \S~\ref{ssec:experimental-setup}, we format our interleaved multimodal reasoning processes as multi-turn conversations. 
To put it simply, an input consists of [\texttt{system prompt}, \texttt{user question}, \texttt{model thought}, \texttt{model tool call}, \texttt{tool output}, \texttt{model thought}, \texttt{model tool call}, \texttt{tool output}, ... ].
However, API specifications of any proprietary models allow this format as is, \ie, either tool outputs cannot include images (OpenAI) or tool outputs need to be included in \texttt{user} messages (Anthropic and Google).
Under this restriction, we, instead, add a note of ``Asking a user to provide tool outputs.'' as tool outputs and add actual tool outputs with images in \texttt{user} messages.
When we spot a case where a model does not generate an answer after $i$ turns or a case where a model generates an answer before $j$ turns, we include a cut-in \texttt{user} message to either encourage the model to answer or use more tools. 
We set the maximum number of turns as $h$, and we stop the iteration regardless of whether or not an answer is generated.
The maximum number of frames that \texttt{sample\_frame} returns is $k$, and the maximum number of frames in an input is $n$.
If more than $k$ frames are selected, we pick $k$ frames at equal intervals. 
If more than $n$ images are included in one prompt, we remove the beginning images until the total number of images is equal to $n$. 
\autoref{fig:tool-definition} contains the detailed definition of our tools in the YAML format (\autoref{fig:tool-definition-text} for the text version). 
At most one tool is executed, even when multiple tool calls are generated. 
When a model outputs multiple tools in one output, we simply pick the first one to execute. 
Hyperparameters are set as follows: $i=5$, $j=2$, $k=32$, $n=64$, $h=10$, $i=8$, and $j=5$. 
Prompt templates are available in~\autoref{fig:prompt-tama-system} and~\ref{fig:prompt-tama-initial}.

\subsection{Model Selection}
\label{appx:model-selection}
Our model selection is mainly based on the performance and capability of a model, under the constraints of our cost budget and academic computational resources.
The following are the reasons for other possible models we did not include in our experiments.
Gemini 2.5 Pro returns server-side errors insufferably frequently at the time we experimented, so we ended up not using it, although its estimated cost is around the same as GPT-5 or Claude 4 Sonnet.
We did not experiment with Claude 4/4.1 Opus due to their high costs.
We did not use Qwen2.5-VL 72B due to the suspicion of its bug related to tool use, more specifically, it outputs a strange character every time it outputs tool calls. 
Following the size of Qwen2.5-VL, we used InternVL3 38B, instead of InternVL3 78B. 
GLM-4.5V~\citep{hong-arxiv2025-glm} was not included because it did not fit into the 4 A6000 GPUs. 
Qwen3-VL~\footnote{\url{https://huggingface.co/Qwen/Qwen3-VL-235B-A22B-Instruct}} came out two days before the deadline of this submission, and we did not include it in our experiments.

The model IDs used in our experiments are as follows: 
\texttt{gpt-5-mini-2025-08-07} (GPT-5 mini), \texttt{gpt-5-2025-08-07} (GPT-5), \texttt{claude-sonnet-4-20250514} (Claude 4 Sonnet), \texttt{gemini-2.5-flash} (Gemini 2.5 Flash), \texttt{Qwen/Qwen2.5-VL-32B-Instruct} (Qwen2.5-VL 32B), \texttt{OpenGVLab/InternVL3-38B} (InternVL3 38B), and \texttt{XiaomiMiMo/MiMo-VL-7B-RL-2508} (MiMo-VL 7B).

\subsection{Other Details}

\begin{table*}[!t]
\centering
\caption{API Cost (USD)}
\fontsize{8}{9}\selectfont
\begin{tabular}{c c c c c}
    \toprule
    Model & Naive & Reasoning & TCoT & TAMA \\
    \midrule
    GPT-5 mini & 1.6 & 0.81 & 5.7 & 4.2 \\
    GPT-5  & 7.6 & 4.8 & 41 & 40 \\
    Claude 4 Sonnet & 10 & 13 & 63 & 41 \\
    Gemini 2.5 Flash & 0.86 & 2.3 & 11 & 5.1 \\
    \bottomrule
\end{tabular}
\label{tab:cost}
\end{table*}

The naive and reasoning approaches receive 32 uniformly sampled frames in their inputs.
API services sometimes show their instability, returning server-side errors. 
In such cases, we run a model one more time to see if we can obtain a result. 
When we do not obtain results after attempting twice, we just include None as an answer. 
To access models, we use APIs for proprietary models and we run locally for open-weight models with the server mode of the \texttt{vllm} library.
For reasoning, we set either ``medium'' or $2048$ for reasoning effort/budget, and $512$ as the maximum number of output tokens. 
Images are all scaled to the resolution of $640\times360$, and we use the center angle for recordings, unless specified.
For Qwen2.5-VL 32B, we used our custom chat template because the original one from the HuggingFace Hub does not contain the templates for tool use. 
For InternVL3 38B, we used a previous version of its chat template because the latest one does not include the templates for tool use. 
For each inference of open-weight models, we used at most 4 A6000 GPUs (48GB memory) throughout our experiments. 
\autoref{tab:cost} shows the reference costs.
Each cost represents the total cost of one model's experiment, \ie, obtaining answers for all questions in the evaluation dataset.

\section{Prompt Template}
\label{appx:prompt}

\begin{figure*}[ht]
    \begin{lstlisting}[style=prompt]
You will be given a question about a video, following frames from the video.
Question: {question}
Return the frame ids which can answer the given question.
Please use the following JSON format for your output:
{
    "frame_ids": [List of integer/frame IDs],
    "justification": "<justification about your output>"
}
    \end{lstlisting}
    \caption{Prompt for frame selection in TCoT.}
    \label{fig:prompt-tcot-frame-selection}
\end{figure*}
\begin{figure*}[ht]
    \begin{lstlisting}[style=prompt]
Frames: {frames}
Parts: {target assembly image}
Instruction: {dot}
An instruction is represented as a directed, acyclic partial graph, where a node is a step and a relation is the order of steps.
For instance, if there is a directed edge between node A and node B (A -> B), A needs to be done before B is performed.
You will be given a question about a video. You are provided frames from the video, retrieved by an intelligent agent. You are also provided with instructions and parts image.
It is crucial that you imagine the visual scene as vividly as possible to enhance the accuracy of your response. Answer in the following format: <answer>your answer</answer>
Question: {question}
    \end{lstlisting}
    \caption{Prompt for answer generation in TCoT.}
    \label{fig:prompt-tcot-answer-generation}
\end{figure*}

\begin{figure*}[ht]
    \begin{lstlisting}[style=prompt]
You are helping a user performing an toy assembly task by checking their activity recording.

You have tools/functions to access the following information:
- video/recording of the activity
- instructions/manuals for the toy
- final picture image of the toy
When you get a question, call the tools to understand the user's current situation so that you can answer the question confidently.
When you finish analyzing the given information, make sure to answer the question in the following format:
<answer>your answer</answer>

Note:
- Each question is asked at the end timing of its recording. So make sure to contextualize each question in the recordings.
- An answer should be one, or a few concise sentence(s).
- Tools can be called multiple times until you obtain enough evidence to answer
the question confidently.
- After each tool call, make sure to think if the returned output is
useful/sufficient for answering the question.
- Each tool can be called multiple times, but tools can be called one at a time.
    \end{lstlisting}
    \caption{System prompt for TAMA}
    \label{fig:prompt-tama-system}
\end{figure*}
\begin{figure*}[ht]
    \begin{lstlisting}[style=prompt]
You are helping a user perform a toy assembly task.

You have tools/functions to access the following information:
- video/recording of the activity in text/caption
- instructions/manuals for the toy in text
- final picture image of the toy in text/caption
When you get a question, call the tools to understand the user's current situation so that you can answer the question confidently.
When you finish analyzing the given information, make sure to answer the question in the following format:
<answer>your answer</answer>

Note:
- Each question is asked at the end timing of its recording. So make sure to contextualize each question in the recordings.
- An answer should be one, or a few concise sentence(s).
- Tools can be called multiple times until you obtain enough evidence to answer
the question confidently.
- After each tool call, make sure to think if the returned output is useful/sufficient for answering the question.
- Each tool can be called multiple times, but tools can be called one at a time.
    \end{lstlisting}
    \caption{System prompt for TAMA (text)}
    \label{fig:prompt-tama-system-text}
\end{figure*}
\begin{figure*}[ht]
    \begin{lstlisting}[style=prompt]
You are helping a user who is performing a toy assembly task by checking their activity recording.

As a starting point, you will be given a question.
Then, you will be given the following information one by one:
- video/recording of the activity
- instructions/manuals for the toy
- target assembly image of the toy
Once you receive all, make sure to answer the question in the following format:
<answer>your answer</answer>

Note:
- Each question is asked at the end of its recording. So make sure to contextualize each question in the recordings.
- An answer should be one or a few concise sentences.
- Make sure to think if the given information is useful/sufficient for answering the question.
    \end{lstlisting}
    \caption{System prompt for our workflow approach in \S~\ref{ssec:agentic-vs-workflow}.}
    \label{fig:prompt-workflow-system}
\end{figure*}
\begin{figure*}[ht]
    \begin{lstlisting}[style=prompt]
I have been working on the task for {duration}.
I have a question. <question>{question}</question>
    \end{lstlisting}
    \caption{Initial user prompt for TAMA}
    \label{fig:prompt-tama-initial}
\end{figure*}

\begin{figure*}[ht]
    \begin{lstlisting}[language=yaml]
- name: "sample_frame"
  description: |
    Function to sample frames in the video between the range with the rate.
    Output consists of a list of 1 fps sampled frame filepaths.
    Frame files are represented with their timestamps in second.
    The maximum number of frames is 30, and if more than the maximum number of frames are requested, the fps rate gets reduced to meet the requirement.
  args:
    start_time: {type: 'string', description: "The start time of the range to sample frames in the format of mm:ss."}
    end_time: {type: 'string', description: "The end time of the range to sample frames in the format of mm:ss."}
    angle: {type: "string", description: "camera angle of the video.", enum: ["center", "top", "right-bottom", "right-center", "right-top", "left-bottom", "left-center", "left-top"]}

- name: "zoom_in"
  description: |
    Function to zoom in one frame.
    You can specify where to zoom-in by a normalized bounding box in the format of [x1,y1,x2,y2], where 0 < x1 < x2 < 1 and 0 < y1 < y2 < 1.
    (x1, y1) corresponds to the top left corner, and (x2,y2) corresponds to the bottom right coner.
  args:
    frame_id: {type: "integer", description: "the id of the frame to zoom-in"}
    angle: {type: 'string', description: "camera angle of the video", enum: ["center", "top", "right-bottom", "right-center", "right-top", "left-bottom", "left-center", "left-top"]}
    bounding_box: {type: "array", description: "normalized bounding box in the format of [x1,y1,x2,y2]", items: {type: "number"}}

- name: "check_instruction"
  description: |
    Function to access the instruction in text or image.
    An instruction is represented as a directed, acycle partial graph, where a node is a step and a relation is a order of steps.
    For instance, if there is a directed edge between node A and node B (A -> B), A needs to be done before B is performed.
    Instructions can be checked in either text or image:
    - text: instructions are represented as text in the DOT format.
    - image: instructions are represented as an figure of a graph.
  args:
    mode: {type: "string", description: "either text or image"}

- name: "check_final_picture"
  description: |
    Function to access the image of the final picture and parts of the target toy car.
    The image may contain its exploded view as well.
  args: null
    \end{lstlisting}
    \caption{TAMA's multimedia-returning tool definitions in the YAML format.}
    \label{fig:tool-definition}
\end{figure*}

\begin{figure*}[ht]
    \begin{lstlisting}[language=yaml]
- name: "sample_frame"
  description: |
    Function that returns a detailed description of sampled frames in the video between a specified range.
    Output consists of one description, based on the sampled frames.
    The default sample rate is 1 fps, and the maximum number of frames is 30.
    If the specified range contains more than 30 frames, i.e., the range exceeds 30 seconds, the fps rate gets reduced so that the number of frames is less than or equal to 30.
  args:
    start_time: {type: 'string', description: "The start time of the range to sample frames in the format of mm:ss."}
    end_time: {type: 'string', description: "The end time of the range to sample frames in the format of mm:ss."}
    angle: {type: "string", description: "camera angle of the video.", enum: ["center", "top", "right-bottom", "right-center", "right-top", "left-bottom", "left-center", "left-top"]}

- name: "zoom_in"
  description: |
    Function that crops one frame and returns the detailed description of the cropped frame.
    You can specify where to zoom-in by a normalized bounding box in the format of [x1,y1,x2,y2], where 0 < x1 < x2 < 1 and 0 < y1 < y2 < 1.
    (x1, y1) corresponds to the top left corner, and (x2,y2) corresponds to the bottom right coner.
    
  args:
    frame_id: {type: "integer", description: "the id of the frame to zoom-in"}
    angle: {type: 'string', description: "camera angle of the video", enum: ["center", "top", "right-bottom", "right-center", "right-top", "left-bottom", "left-center", "left-top"]}
    bounding_box: {type: "array", description: "normalized bounding box in the format of [x1,y1,x2,y2]", items: {type: "number"}}

- name: "check_instruction"
  description: |
    Function that returns the assembly instruction in text.
    An instruction is represented as a directed, acycle partial graph, where a node is a step and a relation is a order of steps.
    For instance, if there is a directed edge between node A and node B (A -> B), A needs to be done before B is performed.
    Instructions are represented as text in the DOT format.
  args: null

- name: "check_final_picture"
  description: |
    Function that returns the detailed description of the final picture, parts image, and possibly with an exploded view as well.
  args: null
    \end{lstlisting}
    \caption{TAMA's text-returning tool definitions in the YAML format.}
    \label{fig:tool-definition-text}
\end{figure*}

\begin{figure*}[ht]
    \begin{lstlisting}[style=prompt]
## Instruction ##
This is an evaluation task.
You will be given a question, gold answer(s), and predicted answer.
Your task is to evaluate if the predicted answer matches against the gold answer(s).

Here is/are the step(s) they have already performed in the actual order:
{previous_steps}

Give your ternary judge 0, 1, or 2:
* 0 means the predicted answer is wrong (unmatch)
* 1 means the predicted answer is partially correct/wrong (partial match)
* 2 means the predicted answer is correct (match)
When multiple gold answers are available (provided as a list), the predicted answer is correct/partially correct if it matches/partially matches with at least one of the gold answers.

Provide your feedback as follows:
## Feedback ##
[Rationale] (your rationale for the judge, as a text)
[Judge] (your judge, as a number, 0, 1, or 2)

## Note ##
The question is being asked by a user who is playing with a take-apart toy.
Gold answer(s) are created by well-trained humans.
Predicted answer is created by a machine, based on the corresponding instruction and the frames of the assemblying process recording.

## Task ##
Now, here are the question, gold answer(s), and predicted answer:
[Question]
{question}
[Gold Answer(s)]
{gold_answer}
[Predicted Answer]
{predicted_answer}

## Feedback ##
[Rationale]
    \end{lstlisting}
    \caption{LLM-as-a-judge prompt}
    \label{fig:prompt-eval}
\end{figure*}

\section{Result}
\label{appx:analysis}

\begin{figure}[ht]
    \centering
    
    \begin{subfigure}[t]{0.3\textwidth}
        \centering
        \includegraphics[width=\textwidth]{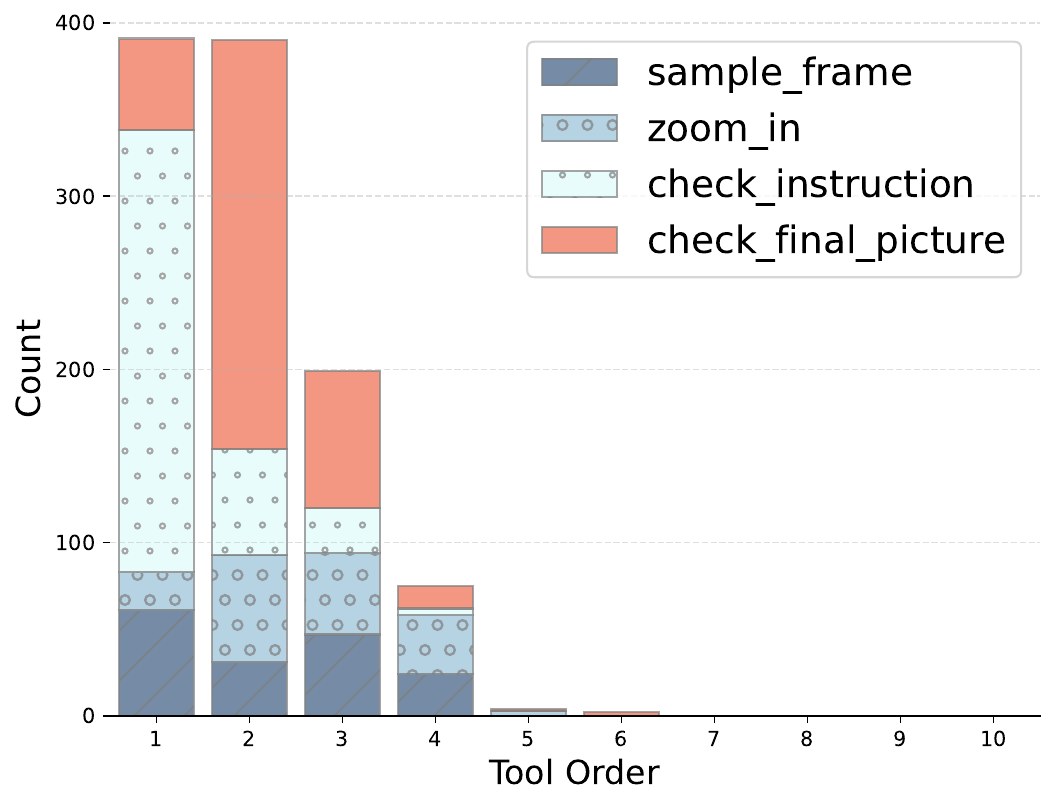}
        \caption{GPT-5 mini}
    \end{subfigure}
    \begin{subfigure}[t]{0.3\textwidth}
        \centering
        \includegraphics[width=\textwidth]{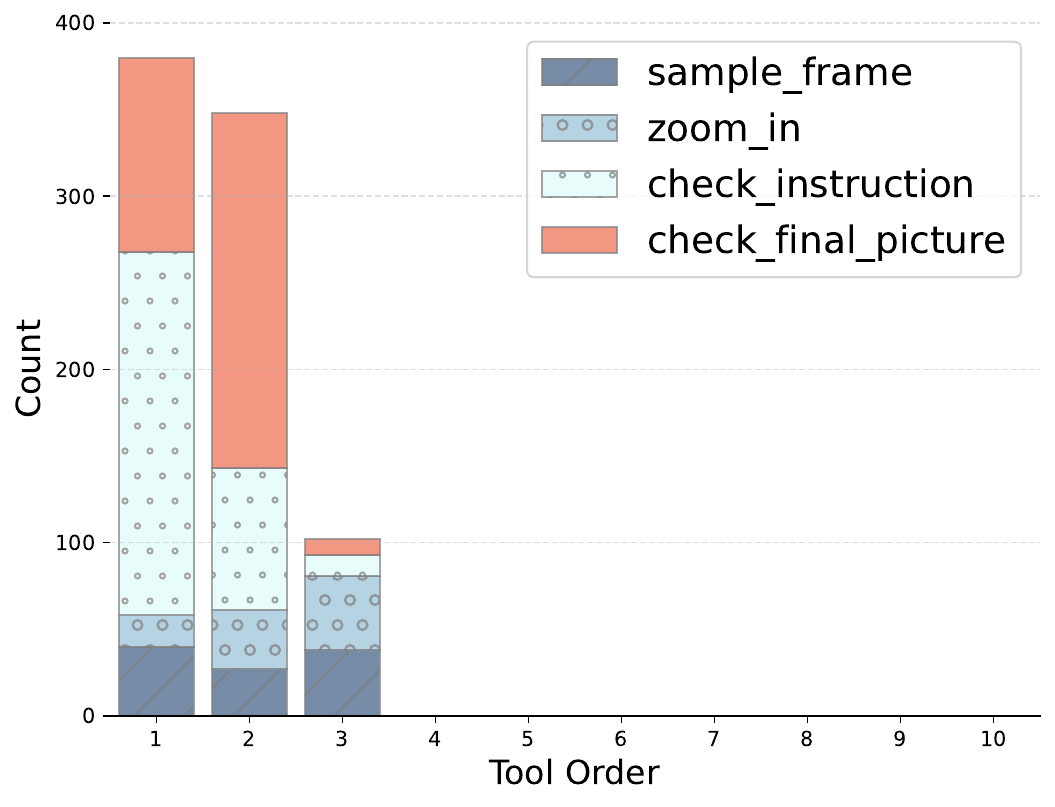}
        \caption{Qwen2.5-VL 32B}
    \end{subfigure}
    \begin{subfigure}[t]{0.3\textwidth}
        \centering
        \includegraphics[width=\textwidth]{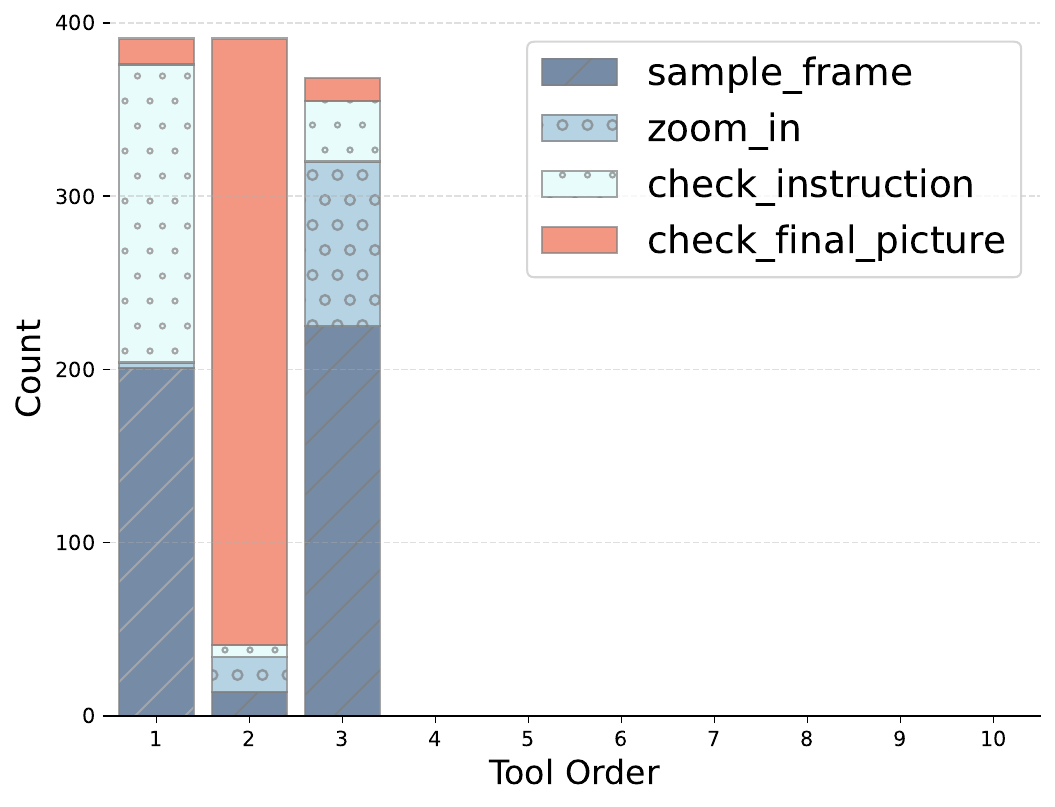}
        \caption{InternVL3 38B}
    \end{subfigure}
    
    \caption{Tool usage pattern for the remaining models.}
    \label{fig:appx-analysis-pattern}
\end{figure}

\autoref{fig:appx-analysis-pattern} shows the remaining models' tool-use patterns.

\section{Limitation}
\label{appx:limitation}

One limitation is in our experiment, regarding model variety. 
While we evaluated both proprietary models and open-weight models, our selection may look small, considering the continuous stream of model releases.
In fact, only a few meet our requirements under our academic computational resources. 
Our framework requires a model to be a VLM that has agentic behavior/tool-use capability. 
Even when the paper/blog of a model mentions the benchmark numbers on agentic tasks, they may not always release ``chat\_template,'' which is crucial to render input information into their specific input format used in their training.
If the templates are not available, we would need to come up with one by educated guesses, which may underrate their capabilities.
Another limitation lies in the cost and efficiency of our framework.
While we observed improved performance for some models, as our framework involves multiple inferences for each question, inference time gets longer with more computational cost, especially compared to the naive approach.
Potential future directions to address this point involve shorter, yet higher-quality multimodal reasoning paths or distillation to smaller models. 
Additionally, the size of the evaluation data may hinder the comparison among models with small differences.

\section{LLM Usage}
We used LLM-powered AI services when drafting this paper, specifically for refining phrases or correcting grammatical errors, but not for ideation or more advanced purposes.

\end{document}